\documentclass{article}

\usepackage[preprint]{neurips_2023}

\usepackage[utf8]{inputenc} 
\usepackage[T1]{fontenc}    
\usepackage{hyperref}       
\usepackage{url}            
\usepackage{booktabs}       
\usepackage{amsfonts}       
\usepackage{nicefrac}       
\usepackage{microtype}      
\usepackage{xcolor}         
\usepackage{amsmath}
\usepackage{multicol}
\usepackage{multirow}
\usepackage{color, colortbl}
\definecolor{Gray}{gray}{0.9}
\usepackage{float}
\usepackage{stfloats}
\newcolumntype{g}{>{\columncolor{Gray}}c}
\usepackage{pifont}
\newcommand{\cmark}{\text{\ding{51}}}
\newcommand{\xmark}{\text{\ding{55}}}
\usepackage{graphicx}
\graphicspath{ {./images/} }

\usepackage{natbib}
\setcitestyle{square,numbers,sort}

\title{{HiTSR}: A Hierarchical Transformer for Reference-based Super-Resolution}

\author{%
  Masoomeh Aslahishahri \\
  Department of Computer Science\\
  University of Saskatchewan\\
  Saskatoon, Canada \\
  \texttt{masi.aslahi@usask.ca} \\
  \And
  Jordan Ubbens \\
  Department of Computer Science \\
  University of Saskatchewan\\
  Saskatoon, Canada \\
  \texttt{jordan.ubbens@usask.ca} \\
  \AND
  Ian Stavness \\
  Department of Computer Science \\
  University of Saskatchewan\\
  Saskatoon, Canada \\
  \texttt{ian.stavness@usask.ca} \\
}

\begin{document}

\maketitle

\begin{abstract}
In this paper, we propose HiTSR, a hierarchical transformer model for reference-based image super-resolution, which enhances low-resolution input images by learning matching correspondences from high-resolution reference images. Diverging from existing multi-network, multi-stage approaches, we streamline the architecture and training pipeline by incorporating the double attention block from GAN literature. Processing two visual streams independently, we fuse self-attention and cross-attention blocks through a gating attention strategy. The model integrates a squeeze-and-excitation module to capture global context from the input images, facilitating long-range spatial interactions within window-based attention blocks. Long skip connections between shallow and deep layers further enhance information flow. Our model demonstrates superior performance across three datasets — SUN80, Urban100, and Manga109. Specifically, on the SUN80 dataset, our model achieves PSNR/SSIM values of 30.24/0.821. These results underscore the effectiveness of attention mechanisms in reference-based image super-resolution. The transformer-based model attains state-of-the-art results without the need for purpose-built subnetworks, knowledge distillation, or multi-stage training, emphasizing the potency of attention in meeting reference-based image super-resolution requirements. 
\end{abstract}

\label{sec:intro}
Image super-resolution (SR) is the process of elevating the spatial resolution of a low-resolution (LR) image to generate a high-resolution (HR) counterpart, effectively recovering intricate textures and details \cite{zheng2018crossnet, zhang2019image}. Image super-resolution can improve the user experience of digital media cont. For example, SR can enhance the texture of video frames in computer games without incurring additional computational costs and enhance the digital zoom functionality of smartphones. Beyond entertainment, applications in computer vision domains like medical imaging and remote sensing can benefit substantially from image super-resolution techniques \cite{li2021review}. The super-resolution challenge is typically categorized into two sub-problems: single image super-resolution (SISR) and reference-based image super-resolution (Ref-SR). Traditional SISR approaches result in blurry images with aliasing artifacts due to significant degradations during downsampling. However, recent SISR models have tried refining the image SR process and creating images resembling their target HR counterparts \cite{dai2019second, kim2016accurate, kim2016deeply, lim2017enhanced, conde2023swin2sr, liang2021swinir}. 

The recent advancements in SR have been primarily driven by Ref-SR, where HR reference images are leveraged to transfer detailed textures to corresponding LR images, resulting in more realistic outputs \cite{yang2020learning}. However, This texture transfer process poses significant challenges, particularly in finding matching correspondences between LR input and HR reference images. Two prominent challenges characterize this task: 1) the spatial (transformation) gap, where similar regions in the LR input and HR reference images exhibit disparities in position and orientation of corresponding landmarks, and 2) the resolution gap, indicating a substantial imbalance in information between LR and HR images due to missing details in the LR input. The spatial gap introduces difficulties in optimal texture transfer, even when the cont is similar, as transformations in scale, position, and shape of objects can lead to discrepancies. Simultaneously, the resolution gap hampers the correspondence matching process, particularly in regions with fine-grained textures, due to the information imbalance between LR and HR images.

Recent progress in Ref-SR has leveraged self-supervised learning \cite{cao2022reference} and incorporated coarse-to-fine matching modules \cite{lu2021masa}, resulting in the improved generation of high-quality HR images. State-of-the-art (SOTA) techniques, exemplified by the $C^2$-Matching architecture \cite{jiang2021robust}, have set the standard for contemporary approaches. This architecture integrates a contrastive correspondence module, trained through student-teacher distillation, to extract correspondences from input and reference images. These correspondences are estimated by a dynamic aggregation module, followed by a restoration module that yields the super-resolved image. While the initial $C^2$-Matching architecture used a convolutional backbone for correspondence extraction, subsequent studies have investigated alternative approaches, including deformable transformers \cite{cao2022reference} or the exploration of a progressive feature alignment and selection module \cite{zhang2022rrsr} to enhance the Ref-SR task further, while retaining the foundational structure of $C^2$-Matching. Despite variations in the correspondence module, these methods typically adhere to the fundamental structure of $C^2$-Matching, incorporating multiple subnetworks and undergoing multiple stages of training. Reproducing the results of $C^2$-Matching and its variants is challenging due to their reliance on a module that does not align with SOTA resources and packages.

This study introduces a novel Ref-SR architecture, HiTSR, to streamline overly complex and multi-stage Ref-SR models, such as $C^2$-Matching, with an end-to-end transformer architecture. Our framework is inspired by the double-attention mechanism \cite{liu2021swin}. HiTSR is designed to jointly learn matching correspondences between two image distributions using self-attention and cross-attention blocks. This approach draws inspiration from the text-to-image domain, where dual streams across two distributions have demonstrated remarkable efficacy in capturing complex semantic relationships \cite{lu2019vilbert}. We enhance our approach by incorporating global attention modules across the LR input and HR reference images. We compute global queries from the input images using a fused residual learning module that includes a squeeze-and-excitation (SE) module, integrating global contextual information to boost the quality of matching correspondences. The self-attention module focuses on information within the LR input, while global query tokens aggregate from the entire HR reference feature map, interacting with key and value representations from the LR image distribution. Long skip connections (LSCs) are employed between shallow and deep layers to facilitate efficient communication across hierarchical levels. In each attention block, we introduce an additional residual learning module, the post-attention residual (PAR) module, positioned before the projection layer to enhance visual quality. Our framework features separate streams for two image distributions, communicating through an adaptive gating attention strategy that regulates attention to self- and cross-attention blocks using a gating parameter. This structure enables the discovery of matching correspondences and interaction between image distributions at varying representation sizes. Ablation studies showcase the significance of each module in the architecture. Unlike existing SOTA methods in Ref-SR, the proposed method features a single end-to-end trainable network, departing from the multi-stage and multi-network architectures used in $C^2$-Matching \cite{jiang2021robust}, DATSR \cite{cao2022reference}, and RRSR \cite{zhang2022rrsr}.
\begin{figure*}[t]
\centering
\includegraphics[width=.99\textwidth]{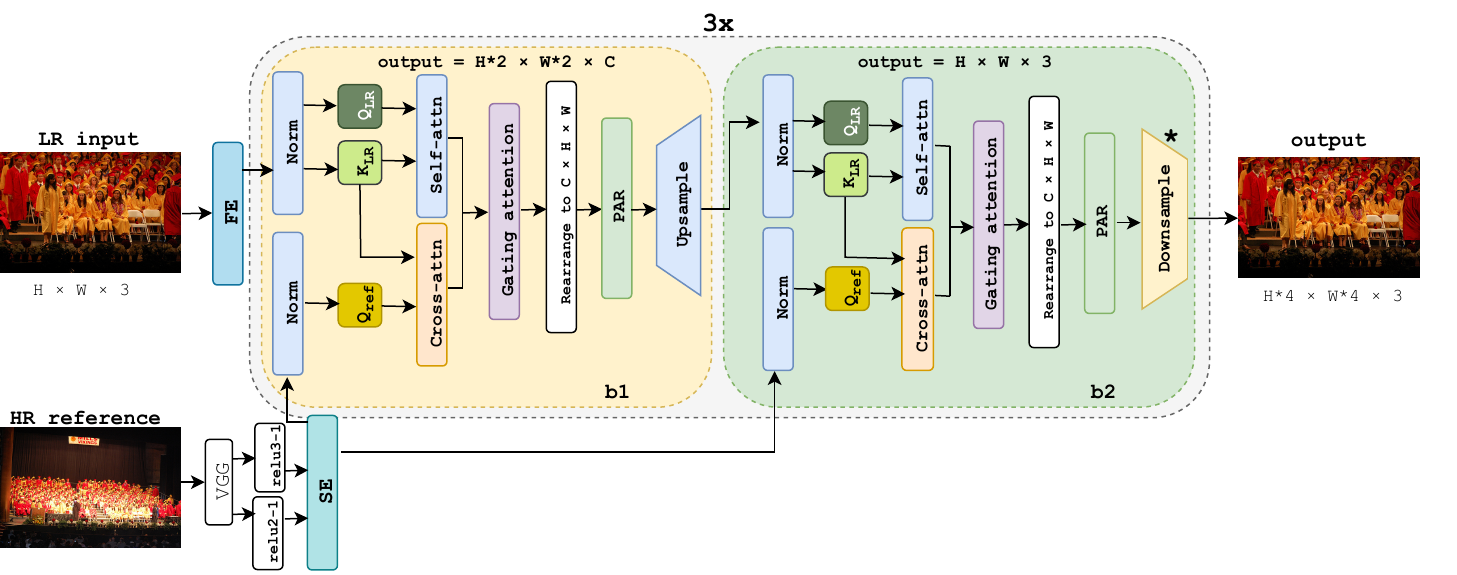}
\caption{Overview of HiTSR. Two input images drive the proposed architecture: LR features, extracted by a deep feature extraction (FE) module containing a 3-block SE layer, and HR reference features, extracted from a pre-trained VGG-based feature extractor. These features undergo further processing through the SE module, with varying depths in layers corresponding to the size of the input image. The sequential arrangement of ``b1" followed by ``b2" blocks incorporates self- and cross-attention modules. In the self-attention module, query and key matrices $Q_{LR}$ and $K_{LR}$ are derived from LR image features. The input query $Q_{ref}$ is sourced from the HR reference image via the SE module for the cross-attention module. Gating attention is applied after the computation of self- and cross-attention matrices within each transformer block. An additional conv-based residual module, termed ``PAR", acts as a post-attention residual block. To traverse various spatial resolutions, we integrate upsampling and downsampling layers. The combination of ``b1" and ``b2" is repeated three times. \\ \textbf{*}: the downsampler module is disabled in the final ``b2" block.}
\label{overview}
\end{figure*}
To the best of our knowledge, our proposed framework represents a novel technique in Ref-SR, achieving quantitative results comparable to other SOTA techniques in the literature but with a much-simplified architecture and training strategy. The main contributions of this study include:
\begin{itemize}
    \item We introduce a novel hierarchical Swin Transformer network incorporating a double attention module to learn joint representations across two image distributions and predict correspondences. This approach enables the model to transfer fine-grained textures from the HR reference image to the corresponding LR image while remaining robust to differences in objects' shape, location, and scale. 
    \item We utilize a CNN-based squeeze and excitation (SE) module to enhance global contextual information, enabling the encoding of spatial features for global query representations across multiple resolutions. Additionally, Long Skip Connections (LSCs) are incorporated between shallow and deep layers of transformer blocks to establish effective communication across hierarchical levels of the network, promoting efficient information flow.
    \item We employ a gating attention strategy to focus on the content of self-attention and cross-attention blocks concurrently. This approach empowers the attention heads to flexibly adjust the feature combination within each transformer block by manipulating a gating parameter. Additionally, an extra convolutional residual learning module (PAR) follows the attention mechanism in each block to enhance the visual quality of the output.
    \item Our framework delivers comparable or superior results to previous methods employing complex training strategies. Across the SUN80, Urban100, and Manga 109 datasets, HiTSR achieves PSNR/SSIM values of 30.2/.819, 32.2/.829, and 20.2/.719 across $l_1$, perceptual, and GAN objectives, surpassing current methods. Moreover, HiTSR achieves competitive PSNR and SSIM values of 30.24/.821, 26.72/.806, and 31.02/.920 solely with the $l_1$ loss function, outperforming all current Ref-SR methods, including those utilizing $C^2$-Matching \cite{jiang2021robust}.
    \end{itemize}

\textbf{Reproducibility.} Code is available at \href{https://github.com/bia006/HiTSR/tree/main?tab=readme-ov-file}{HiTSR}.

\section{Related Work}
\label{sec:related_work}
\textbf{Single Image Super Resolution.}
SISR is a common problem in computer vision and has been studied for an extended period. SISR aims to enhance the spatial details within the downsampled LR image and super-resolve the LR image to the corresponding HR target image. The first deep learning-based SISR study used an interpolated LR image with three-layer CNN to represent the mapping function between the LR and target HR images \cite{dong2015image}. They later improved the SISR process by using a deconvolutional layer to upsample the feature map to the desired size \cite{dong2016accelerating}. Later, deeper networks employing residual learning and dense skip connections were proposed for image SR \cite{kim2016accurate, kim2016deeply, tong2017image, lim2017enhanced, liu2020residual}. The checkerboard artifacts caused by the deconvolutional layer were reduced by replacing it with subpixel convolutional layers to upsample the feature map size \cite{shi2016real}. To improve the performance of SISR models, a channel attention block was proposed to explore inter-channel correlations \cite{zhang2018image}. Different research adapted non-local attention to model long-range dependencies to reproduce quality SR images \cite{dai2019second, liu2018non, zhang2019residual, mei2020image, mei2021image}. Perceptual loss and MSE were introduced to overcome the overly smoothed textures in PSNR-oriented methods \cite{johnson2016perceptual, simonyan2014very}. Generative adversarial networks (GAN) \cite{goodfellow2020generative} were used in image SR models \cite{ledig2017photo} and witnessed further refinement in other studies \cite{sajjadi2017enhancenet, wang2018esrgan, zhang2019ranksrgan}. With the advent of vision transformers, deeper neural networks attending to attention mechanisms have been recently introduced to produce higher quality images while preserving textures \cite{conde2023swin2sr, liang2021swinir, chen2023activating, zhang2022accurate, zhang2022swinfir}. Furthermore, a recent development involves the proposal of a text-mining network to guide image super-resolution for the restoration of high-frequency details \cite{li2024estgn}.

\textbf{Reference-based Image Super Resolution.}
The most significant difference between the SISR and Ref-SR tasks is that the Ref-SR model receives additional information for image SR in the form of an HR reference image. The additional HR reference image and the LR input image can improve the quality of the reconstructed images. The texture representations significantly missing in the LR input image can be transferred from the corresponding HR reference image, which contains the same or similar content. In \cite{zhang2019image}, a multi-scale feature transformation was proposed to fuse swapped features obtained from local patch matching into the LR input image. A coarse-to-fine matching scheme was introduced to reduce the computational complexity while boosting the spatial feature matching when there is a domain shift problem between the LR input and HR reference images \cite{lu2021masa}. In the literature, a contrastive learning network matched relevant correspondences between the LR input and HR reference images \cite{jiang2021robust}. 

$C^2$-Matching is a SOTA model for Ref-SR \cite{jiang2021robust}. The $C^2$-Matching process consists of two main training stages -- one to learn correspondence matching and a second to intake correspondences and synthesize a super-resolved output image. The first stage consists of a contrastive learning module with teacher-student distillation, while the second stage uses a dynamic aggregation module and a restoration network. In the original work, all of the subnetworks are based on CNNs. Other SOTA models have been introduced, borrowing significantly from the structure of $C^2$-Matching while making changes to improve its feature matching capabilities. In \cite{cao2022reference}, a deformable attention block built on UNet \cite{ronneberger2015u} was introduced to attend to the content of feature encoders. A refinement module was also proposed to select/align features for better performance \cite{zhang2022rrsr}. 

Diffusion models have emerged as valuable tools in Ref-SR. A recent approach utilizes a diffusion model conditioned on the unchanged contents of the reference image. This method aims to enhance the content in an LR image, improve the faithfulness of content reconstruction, and enhance the effectiveness of texture transfer, particularly in scenarios involving large scaling factors \cite{dong2024building}.
\section{Approach}
\label{sec:approach}
This study introduces a novel transformer-based architecture named HiTSR, depicted in Figure \ref{overview}. We benchmark our model structure against $C^2$-Matching and designate it as the baseline because our objective is to propose a simpler model than $C^2$-Matching. Our network employs a hierarchical structure to capture multi-scale features efficiently, enabling enhanced understanding and synthesis of complex visual content. Self- and cross-attention mechanisms are integrated into each transformer block to facilitate feature refinement and contextual understanding to capture intricate relationships within and across input data.

Although integrating cross-attention is a common practice in multi-modal domains like text-to-image synthesis \cite{lu2019vilbert}, its application in various vision tasks across diverse distributions of visual data underscores its efficacy in enabling robust feature extraction and knowledge transfer \cite{yang2020learning}. In our methodology, we refer to the input LR image, its corresponding HR reference, the reconstructed output, and the target HR image as $I_{LR}$, $I_{ref}$, $I_{SR}$, and $I_{HR}$, respectively. Each image distribution comprises $N$ samples. 

The HiTSR architecture is designed based on a hierarchical framework incorporating up- and down-sampling blocks, establishing an error feedback mechanism at each stage. The model inputs the LR image and its corresponding reference image, predicting subsequent feature maps for successive stages. Tokens for the transformer blocks are derived using a Swin transformer-inspired patch-based approach, dividing input images into non-overlapping patches and treating these patches from the two distributions as tokens.

Inspired by the concept of double attention in image generation \cite{zhang2022styleswin}, we adapt this technique for Ref-SR. Unlike its original application that involved computing self-attention within local and shifted windows for a single input to broaden the network's receptive field, our application employs double attention across the LR input and HR reference image distributions to conduct self- and cross-attention. This modification allows us to transfer matching correspondences from the HR reference images to their corresponding LR inputs, enhancing the quality of our super-resolution results. Our approach employs a gating attention strategy, effectively combining the content of self-attention and cross-attention blocks simultaneously.
\begin{figure}[tb!]
\centering
\begin{tabular}{cc}
     \includegraphics[scale=.9]{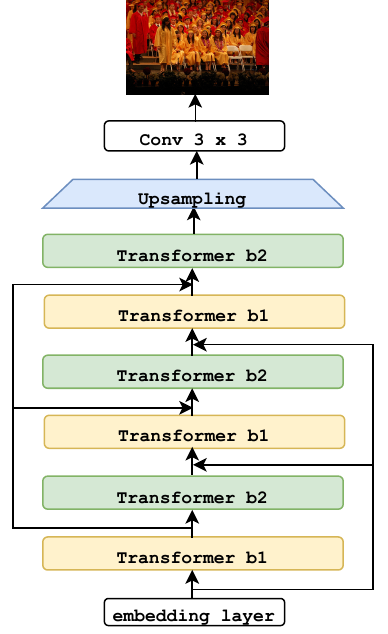} & \includegraphics[scale=.85]{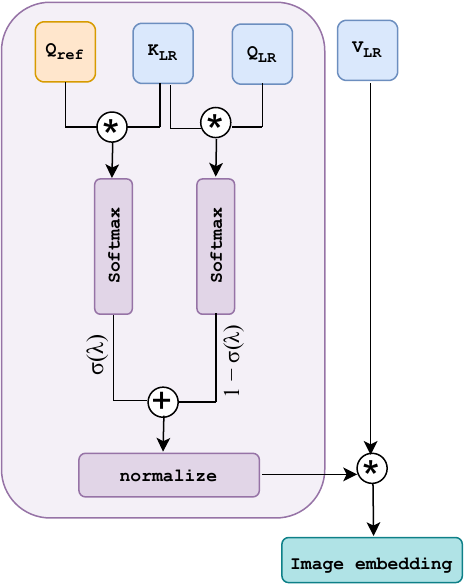}\\
     (a) & (b) \\ 
\end{tabular}
\caption{a) Hierarchical transformer blocks are connected via LSCs between shallow and deep layers, using concatenation to link layers with matching dimensions. The last transformer block (the $3^{rd}$ $b2$) does not include the downsampler module. b) Gating attention balances self- and cross-attention matrices within each transformer block, modulated by a gating parameter $\lambda$.}
\label{gating_attn}
\end{figure}

Furthermore, we integrate LSCs between distinct up- and down-sampling blocks within our framework for enhanced training. These connections create shortcuts between shallow and deep layers, facilitating the training process for image reconstruction. At the final stage, we employ an upsampling block instead of a window-based attention mechanism. This strategy aims to mitigate potential artifacts associated with window-based attention methods.
\subsection{Network Architecture}
We design a hierarchical network containing iterative up- and down-sampling blocks, facilitating the extraction of multi-scale features and enhancing the model's ability to capture intricate details across different resolutions. To facilitate information flow, we incorporate LSCs between iterative blocks. These connections involve concatenating embeddings from previous stages and feeding the combined information into the next transformer block, as depicted in Figure \ref{gating_attn}(a). 

Moreover, we introduce the PAR module, an additional residual learning block positioned after the attention mechanism within each transformer block. The PAR module is positioned before the projection layer and includes a $3 \times 3$ convolutional block. To accommodate this, the 1D sequence of token embeddings $h \in \mathbb{R}^{L \times D}$ is rearranged into a 2D feature of shape $H/P \times W/P \times D$, where $P$ represents the patch size. By integrating the PAR module into the architecture, the model effectively captures and preserves important image details during reconstruction, thereby enhancing overall visual fidelity. Detailed impacts of dropping LSCs and the PAR block are provided in Table \ref{ablation_study}.

In training HiTSR, LR input images ($x \in \mathbb{R}^{H \times W \times 3}$), coupled with their corresponding reference images, serve as input data for the transformer model. The dense features extracted from the reference image, which originate from VGG's $\mathrm{relu3-1}$ and $\mathrm{relu2-1}$ layers, are aligned with the dimensions of the transformer blocks. Inspired by \cite{hatamizadeh2023global} to enhance the significance of input image features, we present a CNN-based spatial feature contraction to improve the discriminative power of the learned representations effectively. A fused conv-based module features a $3 \times 3$ convolutional layer, GELU activation, and an SE \cite{hu2018squeeze} module. The SE improves channel-wise feature representations, emphasizing important features and suppressing less relevant ones. Information integration is achieved through a point-wise convolution, contributing to the module's overall expressiveness. For the LR input image's feature extraction, a deep 3-block SE module is employed, while the depth of feature extraction for the reference images varies by the feature map size.

In our study, the SE block incorporates a squeeze operation, capturing global information through global average pooling across the spatial dimension of the feature map. This operation condenses channel-wise data into a single value per channel, offering a high-level summary of each channel's importance. Subsequently, the excitation operation follows, wherein the condensed information is processed through a small neural network comprised of two fully connected layers with a GELU activation function in between. This network is a gating mechanism, learning to assign weights to each channel based on relevance. The recalibrated channel-wise information is used to reweight the original feature map. Channels deemed more informative receive higher weights, amplifying their influence on the final representation, while less relevant channels are down-weighted. The SE module is ablated in Table \ref{ablation_study}. 
\subsection{Attention Mechanism}
Local attention is confined to querying patches within a local window, which poses challenges when handling variations in object locations, shapes, and scales between LR input and HR reference images. We leverage features extracted from the SE block across the input images to address this limitation and query diverse image regions while operating within the local window. This global representation is attained using global matrices across the LR input and HR reference image distributions, enhancing interaction between the spatial features extracted from the SE block. This approach allows for more flexibility in capturing matching correspondences between LR and HR images with varying object characteristics. 

To compute attention, we adopt a local and shifted window partitioning strategy based on the Swin Transformer architecture, applied to the two input patches of $(I_{LR}, I_{ref})$ within each transformer block. This partitioning enlarges the receptive field by $2.5k$ across each input image, where $k \times k$ represents the window size (i.e., $k = 8$) \cite{zhang2022styleswin}. This approach allows the network to attend to the context of four windows simultaneously—two local and two shifted windows—containing global representations from $I_{LR}$ and global queries from $I_{ref}$. The interaction with richer contextual information embedded in the global matrices, derived from the two input images, facilitates attention to various regions in the input feature maps within each transformer block. The network alternates between the local and shifted windows across the two image distributions to capture essential contextual information. Non-overlapping patches under the local and shifted window partitioning are represented as $I_w$ and $I_{sw}$, respectively as: 

\begin{equation}
    \small
    head_i =
    \begin{cases}
    \text{Self-Attn}(I_{LR_w}W_i^Q, I_{LR_w} W_i^K, I_{LR_w} W_i^V), \\
    \text{Cross-Attn}(I_{ref_{w}} W_i^Q, I_{LR_{w}} W_i^K, I_{LR_{w}} W_i^V)\\
    \text{Self-Attn}(I_{LR_{sw}}W_i^Q, I_{LR_{sw}} W_i^K, I_{LR_{sw}} W_i^V), \\
    \text{Cross-Attn}(I_{ref_{sw}} W_i^Q, I_{LR_{sw}} W_i^K, I_{LR_{sw}} W_i^V)\\ 
    \end{cases}
\end{equation}
where $W_i^Q, W_i^K, W_i^V \in \mathcal{R}^{C \times C}$ are the query, key and value projection matrices for $i^{th}$ head respectively. 

The primary architecture is comprised of iterative Swin Transformer blocks ($i$), managing intermediate visual representations $I_{LR}^i$ and $I_{ref}^i$, initially sized as $(40 \times 40)$ images during training. In each iteration, a transformer block processes the global representations of the LR input $I_{LR}^i$ and the global query of $I_{ref}^i$, extracted from VGG's layer $relu3-1$, yielding weighted feature vectors for the subsequent transformer block. An upsampling module increases the output feature map size to $(80 \times 80)$. Subsequently, another transformer block operates across the generated image embedding from the preceding block and the global representations of $I_{ref}^i$ from VGG's layer $relu2-1$. A CNN-based downsampler, omitted from the final $b2$ block, downsamples the output feature map to $(40 \times 40)$, repeating this process twice more to yield a network with six transformer blocks. LSCs concatenate the feature maps of the same size and use them as input for the next transformer block, as illustrated in Figure \ref{gating_attn}(a). This technique efficiently increases the depth of the network, enhancing its capacity to capture fine-grained details during training. 

In each transformer block of the iterative network, we compute two attention matrices containing self- and cross-attention scores. For self-attention, the global query and key vectors extracted from the $I_{LR}^i$ are passed as inputs to a multi-head attention block where the attention block produces weighted feature vectors for $I_{LR}^i$ conditioned on itself -- in effect attending to different parts of the same input data in a visual stream in each transformer block. For cross-attention, the global query vector extracted from the $I_{ref}^i$ distribution and the key vector extracted from the $I_{LR}^i$ distribution is input to a multi-head attention block, where the attention block produces weighted feature vectors for $I_{LR}^{i}$ conditioned on $I_{ref}^{i}$ -- in effect attending to different parts of reference input data interacting with the LR input data in a visual stream in each transformer block, with residual learning with the initial $I_{LR}^{i}$ representations. 

A gating attention block is strategically employed to mitigate potential differences in magnitudes between self- and cross-attention scores, preventing one attention mechanism from overpowering the other. This gating mechanism regulates the flow of features by dynamically adjusting the influence of self- and cross-attention components during the learning process. By ensuring a balanced contribution from each attention mechanism, the network achieves enhanced stability and robustness in capturing relevant features. 

The Swin block enhances the network's scalability to higher-resolution images, with computational weight scaling linearly with image size. The final block of the proposed framework utilizes an upsampling module, employing pixel shuffling to upscale a feature map from $(80 \times 80)$ to $(160 \times 160)$. Consequently, the framework increases the spatial resolution of the LR input image fourfold.
\subsection{Gating Attention} 
Building upon the insights from \cite{d2021convit}, we employ a gating attention strategy to concurrently combine the content of self-attention and cross-attention blocks, as depicted in Figure \ref{gating_attn}(b). The gating block is initialized with a learnable $\lambda_h$ gating parameter for each attention head $h$, regulating the attention distribution between self-attention and cross-attention scores. The gating attention layer combines the content of self-attention and cross-attention \textit{after} applying softmax, using the gating parameter $\lambda_h$ for each attention head. A sigmoid function maintains the gating parameter in the distribution space. The attention matrix $A^h$ is defined:

\begin{equation}
\begin{split}
    \text{self-attn} &\;=\; \left[ \text{softmax} \left(\frac{Q_{LR}K_{LR}^T}{\sqrt{D_h}}\right) \right] \quad , \\
    \text{cross-attn} &\;=\;  \left[ \text{softmax} \left(\frac{Q_{ref}K_{LR}^T}{\sqrt{D_h}}\right) \right] \quad , \\
    A^h \;=\;  &(1 - \sigma(\lambda_h)) \; \text{self-attn} + \sigma(\lambda_h) \; \text{cross-attn} 
\end{split}
\end{equation}  

\noindent where $ \left[ \; \cdot \; \right]$ is a normalization operation and $\sigma$ denotes the sigmoid function. We initialize the gating parameter $\lambda_h$ to 1 as suggested in \cite{d2021convit}.

The gating strategy dynamically adjusts the importance of each attention score, with each attention head projecting the most relevant representation from the self-attention and cross-attention blocks. The final output is represented as:
\begin{equation}
    \text{attn} = \text{concat}(head_0, ..., head_h)W^O
\end{equation}
\noindent where $W^O \in \mathcal{R}^{C \times C}$ denotes the projection matrix from concatenated heads to output. 

\textbf{Positional Encoding.} Our model incorporates both local and global positional encodings. Local positional encoding involves using relative positional encoding (RPE), capturing spatial information through learnable parameters interacting with queries and keys in each attention block \cite{liu2021swin}. For global positional encoding, we adopt sinusoidal position encoding (SPE), which encodes positional information using a combination of sine and cosine functions \cite{choi2021toward, vaswani2017attention, xu2021positional}. RPE is applied within each transformer block, while SPE is applied to each upsampling block following a transformer block, providing translation invariance and information about the global position.

\textbf{Architectural Details.} Our framework processes input images of size $M \times M$ by dividing them into non-overlapping patches of size $k \times k$ pixels, resulting in $M/k \times M/k$ patches. These patches are embedded into vectors of dimension $D_{emb} = 64N_h$, where $M \times M$, $k \times k$, and $N_h$ represent the input feature map size, window size, and the number of attention heads, respectively. The embedded patches are propagated through iterative up- and down-sampling transformer blocks, maintaining a constant dimensionality. Each block incorporates a double attention mechanism involving $I_{LR}$ and $I_{ref}$, encompassing self-attention and cross-attention operations. Gating attention is applied following the attention mechanism, and a conv-based residual learning module (PAR) is used after the gating attention. The output is passed through a 2-layer MLP with GeLU activation and a residual connection from the initial LR input.

\subsection{Objective Functions}
The objective functions of the proposed method include:\\
\textbf{Reconstruction Loss.} We employ $l_1$ reconstruction loss with a $\text{weight} = 1.0$ for optimization, defined as:
\begin{equation}
    L_{rec} = ||I_{HR} - I_{SR}||_1
\end{equation}
\textbf{Perceptual Loss.} We employ perceptual loss \cite{johnson2016perceptual} to enhance the visual quality of the reconstructed images, defined as:
\begin{equation}
    L_{per} = \frac{1}{V}\sum_{i=1}^C||\Phi_i(I_{HR}) - \Phi_i(I_{SR})||_F 
\end{equation}
\noindent where $C$ represents the channel number, $V$ denotes the volume of the feature maps, and $\Phi$ represents the $relu5-1$ features obtained from the VGG19 network \cite{simonyan2014very}. $||.||_F$ denotes the Frobenius norm. We assign a weight of $1e-4$ to $L_{per}$ for optimization.

\textbf{Adversarial Loss.} We incorporate adversarial training using a multi-scale discriminator architecture inspired by pix2pixHD \cite{wang2018high}. The adversarial loss function employs the hinge loss term \cite{lim2017geometric} with an added $R_1$ gradient penalty for enhanced stability \cite{karras2020analyzing}. The weight assigned to the adversarial loss is set to $1e-6$.
\section{Experiments}
\label{sec:experiments}
\subsection{Experimental Settings}
\textbf{Datasets}. We conduct the performance assessment on various datasets, including CUFED5 \cite{zhang2019image}, Webly-Referenced SR \cite{jiang2021robust},  SUN80 \cite{sun2012super}, Urban100 \cite{huang2015single}, and Manga109 \cite{matsui2017sketch}. For CUFED5, the training set encompasses $11,871$ pairs of input and reference images, while the test set comprises $126$ images. Each input image in the test set is associated with five reference images chosen for their varying degrees of similarity derived from the same event album. The Webly-Referenced SR dataset consists of $80$ image pairs, each comprising an input and a reference image. The SUN80 dataset includes $80$ images, with each input image having $20$ reference images. The evaluation of the Urban100 and Manga109 datasets, which are SISR datasets, follows the strategy outlined in \cite{yang2020learning, zhang2019image}. These datasets contain $100$ and $109$ images, respectively, and a randomly selected image from the same dataset is used as the reference image. The LR input images are generated by downsampling the HR target images by a factor of $4\times$ using bicubic interpolation.

\textbf{Evaluation Metrics}. Peak-Signal-to-Noise Ratio (PSNR) and Structural Similarity Index (SSIM) \cite{wang2004image} are used as evaluation metrics. The Y channel of YCrCb colorspace is used for measuring the PSNR and SSIM values on each dataset. 
\begin{table}[ht!]
\footnotesize
\centering
\resizebox{\textwidth}{!}{ 
\begin{tabular}{ c|c|c|c } 
\toprule
\textbf{\#} & \textbf{Feature extractor module} & \textbf{PAR module} &  \textbf{SE module} \\
\midrule
0 & Conv(64, 64), kernel $(3 \times 3)$, GeLU & rearrange $(input \rightarrow B C H W)$ & AvgPool2d \\ \midrule
1 & SE & Conv(64, 64), kernel = $3 \times 3$, ReLU & Linear (64, 64), GeLU \\ \midrule
2 & Conv (64, 64), kernel $(1 \times 1)$  & Conv(64, 64), kernel = $3 \times 3$, ReLU & Linear (64, 64), Sigmoid \\ \midrule
3 &  Element-wise Addition (input of \#0, output of \#2) & Element-wise Addition (input of \#1, output of \#2) & Element-wise Multiplication (input, \#2) \\ \midrule
4 & Conv(64, 64), kernel $(3 \times 3)$, GeLU & Conv(64, 64), kernel = $3 \times 3$, ReLU & - \\ \midrule
5 & SE & Conv(64, 64), kernel = $3 \times 3$, ReLU &  - \\ \midrule
6 & Conv (64, 64), kernel $(1 \times 1)$ & Element-wise Addition (input of \#4, output of \#5) &  - \\ \midrule
7 &  Element-wise Addition(input of \#4, output of \#6) & rearrange $(\#6 \rightarrow B, H*W, C)$ & - \\ \midrule
8 & Conv(64, 64), kernel $(3 \times 3)$, GeLU & Linear (64, 64), GeLU & - \\ \midrule
9 & SE & - & - \\ \midrule
10 & Conv (64, 64), kernel $(1 \times 1)$  & - & - \\ \midrule
11 &  Element-wise Addition (input of \#8, output of \#10) & - & - \\ 
\bottomrule
& \textbf{(a)} & \textbf{(b)} & \textbf{(c)} \\
\end{tabular}
} 
\caption{Distinct components of our framework: (a) depicts the feature extractor structure applied to both the input LR image and the extracted VGG features of the lowest resolution of HR reference images, (b) illustrates the architecture of the PAR module incorporated into the attention block, and (c) showcases the structure of the SE module implemented in the feature extraction module.}
\label{module_table}
\end{table}

\subsection{Implementation Details}
The training process employs the Adam solver \cite{kingma2014adam} with $\beta_1=0.9$ and $\beta_2=0.999$. A MultiStep learning rate schedule is adopted, setting the initial learning rate to $2e-4$. Data augmentation includes vertical, horizontal, and random rotations to enlarge the dataset. The training is performed with a batch size of nine. A downsampled input LR image with size $(256 \times 256)$ is utilized during inference. Corresponding patches are extracted from HR reference images using $relu3 - 1$ and $relu2 - 1$ layers of a pre-trained VGG network. Distinct components of the networks, namely the feature extraction module and the SE and PAR modules, are elaborated on in the subsequent subsection. 

\subsubsection{Network Structures}
\label{sec:Supplementary_A}
\textbf{Feature Extractor incorporating SE Module}.
Table \ref{module_table}-(a) provides a comprehensive overview of the feature extractor structure, encompassing the SE module. Given the varying resolutions of the HR reference images, we employ the feature extractor module with two distinct depths. Table \ref{module_table}-(a) outlines the architecture of the deeper extractor module employed for the LR input image and the lowest resolution in the HR reference image. 

\textbf{Post-Attention Residual (PAR) Module}.
Table \ref{module_table}-(b) outlines the Post-Attention Residual (PAR) module, featuring a residual connection after the attention layer. This design enhances the model's ability to preserve and propagate essential information effectively. The PAR module employs convolution-based residual learning, necessitating tensor rearrangement for optimal implementation. 

\textbf{Squeeze and Excitation (SE) Module}.
Table \ref{module_table}-(c) overviews the SE module, which utilizes average pooling and convolutional operations. These operations work synergistically to selectively highlight informative features and suppress less relevant ones within various channels of a convolutional layer.
\begin{table*}[t!]
\small
\centering
\begin{tabular}{c| l c c c c c} 
 \hline
 & Method & CUFED5 & SUN80 & Urban100 & Manga109 & WR-SR \\ 
 \hline\hline
 & SRCNN \cite{dong2015image} & 25.33 / .745 & 28.26 / .781 & 24.41 / .738 & 27.2 / .850 & 27.27 / .767\\ 
 & EDSR \cite{lim2017enhanced} & 25.93 / .777 & 28.52 / .792 & 25.51 / .783 & 28.93 / .891 & 28.07 / .793 \\
 SISR & RCAN \cite{zhang2018image} & 26.06 / .769 & 29.86 / .810 & 25.42 / .768 & 29.38 / .895 & 28.25 / .799\\
\rowcolor{lightgray}
\cellcolor{white} & ESRGAN \cite{wang2018esrgan}  & 21.90 / .633 & 24.18 / .651 & 20.91 / .620 & 23.53 / .797 & 26.07 / .726 \\ 
& ENet \cite{sajjadi2017enhancenet} & 24.24 / .695 & 26.24 / .702 & 23.63 / .711 &  25.25 / .802 & 25.47 / .699 \\
 \rowcolor{lightgray}
\cellcolor{white} & RankSRGAN \cite{zhang2019ranksrgan} & 22.31 / .635 & 25.60 / .667 & 21.47 / .624 & 25.04 / .803 & 26.15 / .719 \\
\hline\hline
& CrossNet \cite{zheng2018crossnet} & 25.48 / .764 & 28.52 / .793 & 25.11 / .764 & 23.36 / .741 & -\\
\rowcolor{lightgray}
\cellcolor{white} &  SRNTT \cite{zhang2019image} \ & 25.61 / .764 & 27.59 / .756 & 25.09 / .774 & 27.54 / .862 & 26.53 / .745\\
& SRNTT-\textit{rec} \cite{zhang2019image} & 26.24 / .784 & 28.54 / .793 & 25.50 / .783 & 28.95 / .885 & 27.59 / .780\\
\rowcolor{lightgray} 
\cellcolor{white} & MASA \cite{lu2021masa} & 24.92 / .729 & 27.12 / .708 & 23.78 / .712 & 27.44 / .849 & 25.76 / .717\\
& MASA-\textit{rec} \cite{lu2021masa} & 27.54 / .814 & 30.15 / .815 & 26.09 / .786 & 30.28 / .909 & 28.19 / .796\\
\rowcolor{lightgray} 
\cellcolor{white} &  TTSR \cite{yang2020learning} & 25.53 / .765 & 28.59 / .774 & 24.62 / .747 & 28.70 / .886 & 26.83 / .762\\
& TTSR-\textit{rec} \cite{yang2020learning} & 27.09 / .804 & 30.02 / .814 & 25.87 / .784 & 30.09 / .907 & 27.97 / .792\\

\rowcolor{lightgray}
\cellcolor{white} Ref-SR &  $C^2$-Matching \cite{jiang2021robust}  & 27.16 / .805 & 29.75 / .799 & 25.52 / .764 & 29.73 / .893 & 27.80 / .780\\
& $C^2$-Matching-\textit{rec} \cite{jiang2021robust} & 28.24 / .841 & 30.18 / .817 & 26.03 / .785 & 30.47 / .911 & 28.32 / .801\\
\rowcolor{lightgray}
\cellcolor{white} &  DATSR \cite{cao2022reference}  & 27.95 / \textcolor{red}{.835} & 29.77 / .800 & 25.92 / .775 & 29.75 / .893 & 27.87 / \textcolor{red}{.787}\\
& DATSR-\textit{rec} \cite{cao2022reference} & 28.72 / \textcolor{blue}{.856} & 30.20 / .818 & 26.52 / .798 & 30.49 / .912 & 28.34 / \textcolor{blue}{.805}\\
\rowcolor{lightgray} 
\cellcolor{white} & RRSR \cite{zhang2022rrsr} & \textcolor{red}{28.09 / .835} & 29.57 / .793 & 25.68 / .767 & 29.82 / .893 & \textcolor{red}{27.89} / .784\\
& RRSR-\textit{rec} \cite{zhang2022rrsr} & \textcolor{blue}{28.83 / .856} & 30.13 / .816 & 26.21 / .790 & 30.91 / .913 & \textcolor{blue}{28.41} / .804\\
\rowcolor{lightgray} 
\cellcolor{white} & HiTSR (ours) & 26.15 / .771 & \textcolor{red}{30.04 / .805}  & \textcolor{red}{25.98 / .783} & \textcolor{red}{30.08/ .903} & 27.86 / .784 \\
& HiTSR-\textit{rec} (ours) & 27.08 / .801 & \textcolor{blue}{30.24 / .821}  & \textcolor{blue}{26.72 / .806} & \textcolor{blue}{31.02 / .918} & 28.26 / .802\\
\hline
\end{tabular}
\caption{Quantitative comparisons using PSNR / SSIM metrics. The SISR and Ref-SR methods are grouped accordingly. The \textit{`-rec'} denotes only reconstruction ($l_1$) loss. The methods shaded in grey use $l_1$ loss, perceptual loss and GAN loss. The highest values for networks trained with multiple loss functions are shown in \textcolor{red}{red}, and the highest values for networks trained with only reconstruction $l_1$ loss are shown in \textcolor{blue}{blue}.}
\label{quantitative_comp}
\end{table*}
\begin{figure*}[tb!]
\centering
\footnotesize
\begin{tabular}{ccccc}
\multicolumn{2}{c}{\includegraphics[width=.25\linewidth,height=1.05cm]{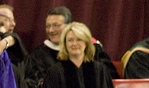}}&
\multicolumn{1}{c}{\includegraphics[width=.15\linewidth,height=1.05cm]{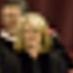}}&
\multicolumn{1}{c}{\includegraphics[width=.15\linewidth,height=1.05cm]{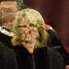}}&
\multicolumn{1}{c}{\includegraphics[width=.15\linewidth,height=1.05cm]{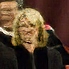}}\\
\multicolumn{2}{c}{target image} & bicubic & ESRGAN & ENet \\
\multicolumn{2}{c}{\includegraphics[width=.25\linewidth,height=1.05cm]{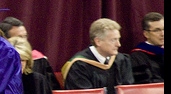}}&
\multicolumn{1}{c}{\includegraphics[width=.15\linewidth,height=1.05cm]{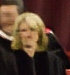}}&
\multicolumn{1}{c}{\includegraphics[width=.15\linewidth,height=1.05cm]{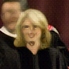}}&
\multicolumn{1}{c}{\includegraphics[width=.15\linewidth,height=1.05cm]{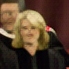}}\\
\multicolumn{2}{c}{reference image} & SRNTT & $C^2$-Matching & HiTSR (ours)\\~\\
\multicolumn{2}{c}{\includegraphics[width=.25\linewidth,height=1.05cm]{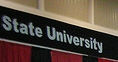}}&
\multicolumn{1}{c}{\includegraphics[width=.15\linewidth,height=1.05cm]{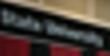}}&
\multicolumn{1}{c}{\includegraphics[width=.15\linewidth,height=1.05cm]{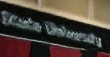}}&
\multicolumn{1}{c}{\includegraphics[width=.15\linewidth,height=1.05cm]{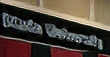}}
\\
\multicolumn{2}{c}{target image} & bicubic & ESRGAN & ENet \\
\multicolumn{2}{c}{\includegraphics[width=.25\linewidth,height=1.05cm]{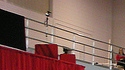}}&
\multicolumn{1}{c}{\includegraphics[width=.15\linewidth,height=1.05cm]{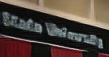}}&
\multicolumn{1}{c}{\includegraphics[width=.15\linewidth,height=1.05cm]{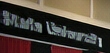}}&
\multicolumn{1}{c}{\includegraphics[width=.15\linewidth,height=1.05cm]{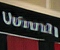}}
\\
\multicolumn{2}{c}{reference image} & SRNTT & $C^2$-Matching & HiTSR (ours)\\~\\
\multicolumn{2}{c}{\includegraphics[width=.25\linewidth,height=1.05cm]{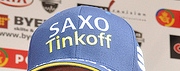}}&
\multicolumn{1}{c}{\includegraphics[width=.15\linewidth,height=1.05cm]{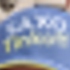}}&
\multicolumn{1}{c}{\includegraphics[width=.15\linewidth,height=1.05cm]{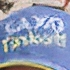}}&
\multicolumn{1}{c}{\includegraphics[width=.15\linewidth,height=1.05cm]{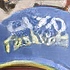}}
\\ 
\multicolumn{2}{c}{target image} & bicubic & ESRGAN & ENet \\
\multicolumn{2}{c}{\includegraphics[width=.25\linewidth,height=1.05cm]{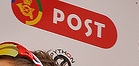}}&
\multicolumn{1}{c}{\includegraphics[width=.15\linewidth,height=1.05cm]{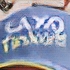}}&
\multicolumn{1}{c}{\includegraphics[width=.15\linewidth,height=1.05cm]{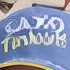}}&
\multicolumn{1}{c}{\includegraphics[width=.15\linewidth,height=1.05cm]{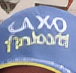}}
\\
\multicolumn{2}{c}{reference image} & SRNTT & $C^2$-Matching & HiTSR (ours)\\~\\
\multicolumn{2}{c}{\includegraphics[width=.25\linewidth,height=1.05cm]{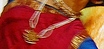}}&
\multicolumn{1}{c}{\includegraphics[width=.15\linewidth,height=1.05cm]{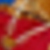}}&
\multicolumn{1}{c}{\includegraphics[width=.15\linewidth,height=1.05cm]{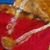}}&
\multicolumn{1}{c}{\includegraphics[width=.15\linewidth,height=1.05cm]{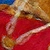}} \\
\multicolumn{2}{c}{target image} & bicubic & ESRGAN & ENet \\
\multicolumn{2}{c}{\includegraphics[width=.25\linewidth,height=1.05cm]{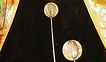}}&
\multicolumn{1}{c}{\includegraphics[width=.15\linewidth,height=1.05cm]{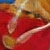}}&
\multicolumn{1}{c}{\includegraphics[width=.15\linewidth,height=1.05cm]{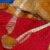}}&
\multicolumn{1}{c}{\includegraphics[width=.15\linewidth,height=1.05cm]{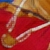}}\\
\multicolumn{2}{c}{reference image} & SRNTT & $C^2$-Matching & HiTSR (ours)\\~\\
\multicolumn{2}{c}{\includegraphics[width=.25\linewidth,height=1.05cm]{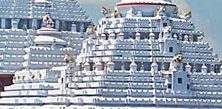}}&
\multicolumn{1}{c}{\includegraphics[width=.15\linewidth,height=1.05cm]{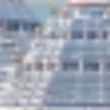}}&
\multicolumn{1}{c}{\includegraphics[width=.15\linewidth,height=1.05cm]{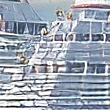}}&
\multicolumn{1}{c}{\includegraphics[width=.15\linewidth,height=1.05cm]{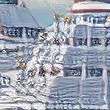}}\\
\multicolumn{2}{c}{target image} & bicubic & ESRGAN & ENet \\
\multicolumn{2}{c}{\includegraphics[width=.25\linewidth,height=1.05cm]{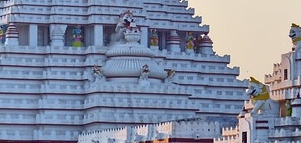}}&
\multicolumn{1}{c}{\includegraphics[width=.15\linewidth,height=1.05cm]{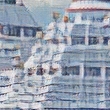}}&
\multicolumn{1}{c}{\includegraphics[width=.15\linewidth,height=1.05cm]{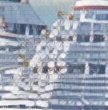}}&
\multicolumn{1}{c}{\includegraphics[width=.15\linewidth,height=1.05cm]{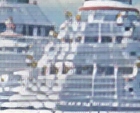}}\\
\multicolumn{2}{c}{reference image} & SRNTT & $C^2$-Matching & HiTSR (ours)\\~\\
\multicolumn{2}{c}{\includegraphics[width=.25\linewidth,height=1.05cm]{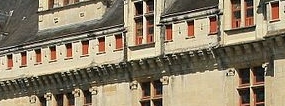}}&
\multicolumn{1}{c}{\includegraphics[width=.15\linewidth,height=1.05cm]{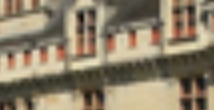}}&
\multicolumn{1}{c}{\includegraphics[width=.15\linewidth,height=1.05cm]{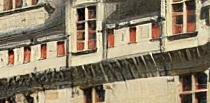}}&
\multicolumn{1}{c}{\includegraphics[width=.15\linewidth,height=1.05cm]{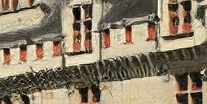}}
\\ 
\multicolumn{2}{c}{target image} & bicubic & ESRGAN & ENet \\
\multicolumn{2}{c}{\includegraphics[width=.25\linewidth,height=1.05cm]{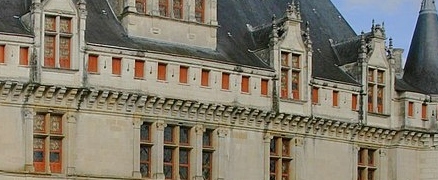}}&
\multicolumn{1}{c}{\includegraphics[width=.15\linewidth,height=1.05cm]{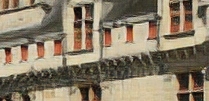}}&
\multicolumn{1}{c}{\includegraphics[width=.15\linewidth,height=1.05cm]{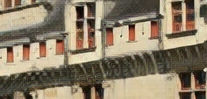}}&
\multicolumn{1}{c}{\includegraphics[width=.15\linewidth,height=1.05cm]{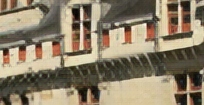}}
\\ 
\multicolumn{2}{c}{reference image} & SRNTT & $C^2$-Matching & HiTSR (ours)\\
\end{tabular}
\caption{Qualitative comparisons, for example, images from CUFED5 and the Webly-Ref-SR datasets. Top row: target image, bicubic, ESRGAN, ENet results. The bottom row: reference image, SRNTT, $C^2$-Matching, and our HiTSR outcomes. All methods utilize $l_1$ loss, perceptual loss, and GAN loss objectives.}
\label{qualitative_results}
\end{figure*}
\subsection{Results and Analysis}
We compare the proposed technique with recent CNN- and transformer-based networks from the literature. For SISR methods, we include SRCNN \cite{dong2015image}, EDSR \cite{lim2017enhanced}, RCAN \cite{zhang2018image}, ESRGAN \cite{wang2018esrgan}, ENet \cite{sajjadi2017enhancenet} and RankSRGAN \cite{zhang2019ranksrgan}. For Ref-SR methods, we include CrossNet \cite{zheng2018crossnet}, SRNTT \cite{zhang2019image}, MASA \cite{lu2021masa}, TTSR \cite{yang2020learning}, $C^2$-Matching \cite{jiang2021robust}, DATSR \cite{cao2022reference} and RRSR \cite{zhang2022rrsr}.
 
\textbf{Quantitative results.} Table \ref{quantitative_comp} shows a quantitative comparison of HiTSR against existing SOTA methods. The proposed method demonstrates competitive results with SOTA across various datasets, outperforming the transformer-based model TTSR \cite{yang2020learning} regarding SSIM or PSNR values. Our model achieves SOTA in the SUN80, Urban100, and Manga109 benchmarks. Techniques using the general $C^2$-Matching architecture \cite{cao2022reference, zhang2022rrsr} still achieve the best performance in the CUFED5 and WR-SR datasets. The observed lower performance of HiTSR on the CUFED5 and WR-SR datasets could be attributed to the diverse distribution of these datasets and the relatively lower complexity of HiTSR in capturing the complex characteristics of their content. However, it has been mentioned that the $C^2$-Matching architecture and its subsequent variants necessitate a more complex architecture and training procedure to attain their superior results.

\textbf{Qualitative results.} Figure \ref{qualitative_results} shows qualitative comparisons with SOTA methods on CUFED5 dataset \cite{zhang2019image} and the Webly-Referenced SR dataset \cite{jiang2021robust}, where each image is paired with a reference image. We compare HiTSR with bicubic, ESRGAN \cite{wang2018esrgan}, ENet \cite{sajjadi2017enhancenet}, SRNTT \cite{zhang2019image} and $C^2$-Matching \cite{jiang2021robust} across all three loss functions: $l_1$ loss, perceptual loss, and GAN loss. The output of HiTSR exhibits fewer artifacts than competing models. In the top example, our model reconstructs the face and neck more accurately than the other SOTA models. In the second and third examples, HiTSR demonstrates improved accuracy in reconstructing alphabetical letters, making words like \textit{``State University"} and \textit{``SAXO"}, with particular emphasis on \textit{X} more readable than competing models. Furthermore, HiTSR successfully reconstructs fine features of buildings and necklaces without mismatching features and their relative positions. Overall, HiTSR preserves the integrity of global geometry while enhancing fine-grained texture reconstruction.

\begin{figure}[H]
\centering
\small
\begin{tabular}{cccc}
input & target image & $C^2$-Matching & HiTSR (ours) \\ 
\includegraphics[width=.21\linewidth,height=1.98cm]{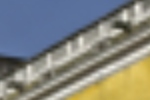}&
\includegraphics[width=.21\linewidth,height=1.98cm]{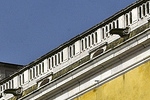}&
\includegraphics[width=.21\linewidth,height=1.98cm]{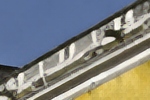}&
\includegraphics[width=.21\linewidth,height=1.98cm]{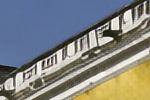} \\
\includegraphics[width=.21\linewidth,height=1.98cm]{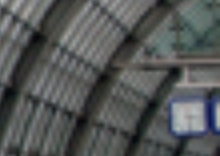}&
\includegraphics[width=.21\linewidth,height=1.98cm]{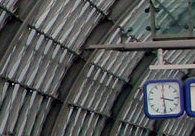}&
\includegraphics[width=.21\linewidth,height=1.98cm]{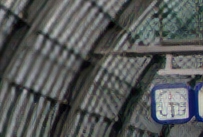}&
\includegraphics[width=.21\linewidth,height=1.98cm]{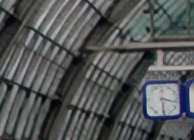} \\ 
\includegraphics[width=.21\linewidth,height=1.98cm]{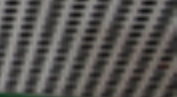}&
\includegraphics[width=.21\linewidth,height=1.98cm]{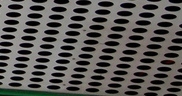}&
\includegraphics[width=.21\linewidth,height=1.98cm]{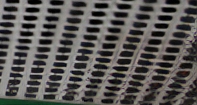}&
\includegraphics[width=.21\linewidth,height=1.98cm]{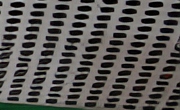} \\
\includegraphics[width=.21\linewidth,height=1.98cm]{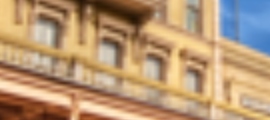}&
\includegraphics[width=.21\linewidth,height=1.98cm]{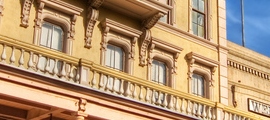}&
\includegraphics[width=.21\linewidth,height=1.98cm]{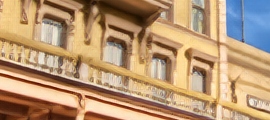}&
\includegraphics[width=.21\linewidth,height=1.98cm]{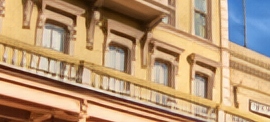} \\
\includegraphics[width=.21\linewidth,height=1.98cm]{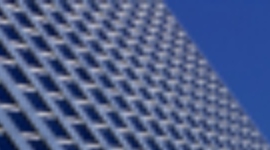}&
\includegraphics[width=.21\linewidth,height=1.98cm]{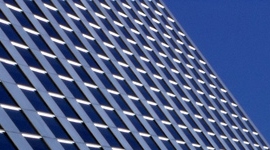}&
\includegraphics[width=.21\linewidth,height=1.98cm]{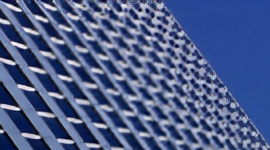}&
\includegraphics[width=.21\linewidth,height=1.98cm]{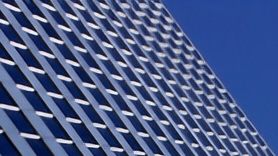} \\ 
\includegraphics[width=.21\linewidth,height=1.98cm]{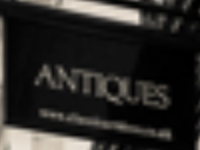}&
\includegraphics[width=.21\linewidth,height=1.98cm]{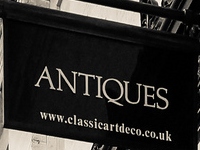}&
\includegraphics[width=.21\linewidth,height=1.98cm]{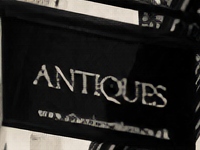}&
\includegraphics[width=.21\linewidth,height=1.98cm]{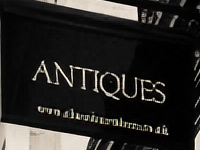} \\
\includegraphics[width=.21\linewidth,height=1.98cm]{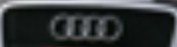}&
\includegraphics[width=.21\linewidth,height=1.98cm]{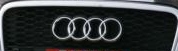}&
\includegraphics[width=.21\linewidth,height=1.98cm]{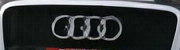}&
\includegraphics[width=.21\linewidth,height=1.98cm]{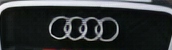} \\ 
\includegraphics[width=.21\linewidth,height=1.98cm]{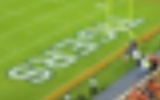}&
\includegraphics[width=.21\linewidth,height=1.98cm]{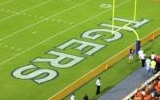}&
\includegraphics[width=.21\linewidth,height=1.98cm]{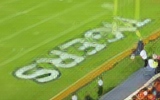}&
\includegraphics[width=.21\linewidth,height=1.98cm]{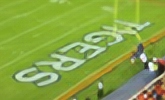} \\
\includegraphics[width=.21\linewidth,height=1.98cm]{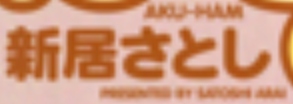}&
\includegraphics[width=.21\linewidth,height=1.98cm]{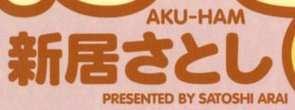}&
\includegraphics[width=.21\linewidth,height=1.98cm]{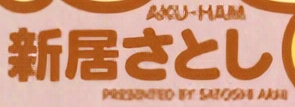}&
\includegraphics[width=.21\linewidth,height=1.98cm]{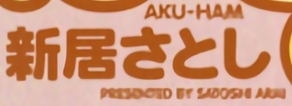} \\
\includegraphics[width=.21\linewidth,height=1.98cm]{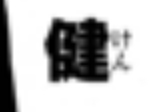}&
\includegraphics[width=.21\linewidth,height=1.98cm]{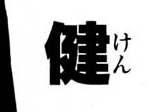}&
\includegraphics[width=.21\linewidth,height=1.98cm]{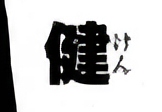}&
\includegraphics[width=.21\linewidth,height=1.98cm]{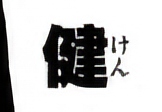} \\
\end{tabular}
\caption{Qualitative comparisons of our results on Urban100, Sun80, and Manga109 datasets with $C^2$-Matching. Each method incorporates $l_1$ loss, perceptual loss, and GAN loss objectives.}
\label{additional_qualitative_results}
\end{figure}

In Figure \ref{additional_qualitative_results}, additional visual comparisons with $C^2$-Matching \cite{jiang2021robust} are showcased, covering other datasets including Urban100 \cite{huang2015single}, SUN80 \cite{sun2012super}, and Manga109 \cite{matsui2017sketch}, where a random image from the same dataset is utilized as a reference image. The images restored by our proposed HiTSR model exhibit competitive or superior visual quality. These results are achieved while maintaining the integrity of the model's objectives, including $l_1$ loss, perceptual loss, and GAN loss.

\begin{table}[htb!]
\scriptsize
\centering
\begin{tabular}{ c|cccccc|c} 
\toprule
 & Self-attn & Cross-attn & SE & LSCs & PAR & Gating-attn & PSNR/SSIM\\
\midrule
a) & \cmark & \cmark & \cmark & \cmark & \cmark & \cmark & \textbf{26.52~/~.780} \\ 
b) & \textcolor{red}{\xmark} & \cmark & \cmark & \cmark & \cmark & \textcolor{red}{\xmark} &  26.11~/~.767\\ 
c) & \cmark & \textcolor{red}{\xmark} & \cmark & \cmark & \cmark & \textcolor{red}{\xmark} & 26.20~/~.772 \\   
d) & \cmark & \cmark & \cmark & \textcolor{red}{\xmark} & \cmark & \cmark & 25.90~/~75.6\\ 
e) & \cmark & \cmark & \cmark & \cmark & \textcolor{red}{\xmark} & \cmark & 26.10~/~.769\\ 
f) & \cmark & \cmark & \textcolor{red}{\xmark} & \cmark & \cmark & \cmark & 25.75~/~.751\\ 
g) & \cmark & \cmark & \textcolor{red}{\xmark} & \textcolor{red}{\xmark} & \cmark & \cmark & 25.70~/~.749\\ 
h) & \cmark & \cmark & \textcolor{red}{\xmark} & \cmark & \textcolor{red}{\xmark} & \cmark & 26.02~/~.765\\ 
i) & \cmark & \cmark & \cmark &  \textcolor{red}{\xmark} & \textcolor{red}{\xmark} & \cmark & 25.82~/~.760\\ 
j) & \cmark & \cmark & \textcolor{red}{\xmark} &  \textcolor{red}{\xmark} & \textcolor{red}{\xmark} & \cmark & 25.50~/~.747\\ 

\bottomrule
\end{tabular}
\caption{Ablation studies were conducted on HiTSR-\textit{rec}, trained for 100K iterations on the CUFED5 dataset. Due to time constraints, ten ablation studies were carried out, omitting various network components to assess their impact on performance in Ref-SR.}
\label{ablation_study}
\end{table}
\begin{figure}[H]
\centering
\small
\begin{tabular}{ccccc}
input & reference image & HR target & ablation (j) & ablation (a) \\ 
\includegraphics[width=.16\linewidth,height=1.75cm]{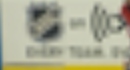}&
\includegraphics[width=.16\linewidth,height=1.75cm]{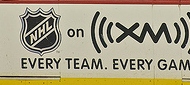}&
\includegraphics[width=.16\linewidth,height=1.75cm]{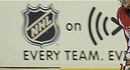}&
\includegraphics[width=.16\linewidth,height=1.75cm]{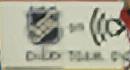} &
\includegraphics[width=.16\linewidth,height=1.75cm]{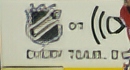} \\ 
\includegraphics[width=.16\linewidth,height=1.75cm]{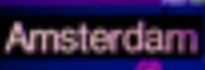}&
\includegraphics[width=.16\linewidth,height=1.75cm]{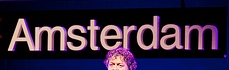}&
\includegraphics[width=.16\linewidth,height=1.75cm]{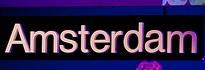}&
\includegraphics[width=.16\linewidth,height=1.75cm]{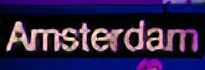} &
\includegraphics[width=.16\linewidth,height=1.75cm]{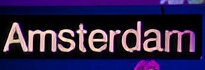} \\ 
\includegraphics[width=.16\linewidth,height=1.75cm]{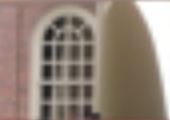}&
\includegraphics[width=.16\linewidth,height=1.75cm]{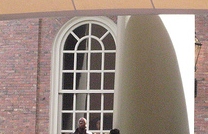}&
\includegraphics[width=.16\linewidth,height=1.75cm]{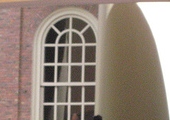}&
\includegraphics[width=.16\linewidth,height=1.75cm]{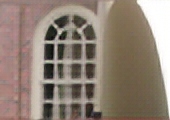} &
\includegraphics[width=.16\linewidth,height=1.75cm]{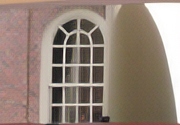} \\ 
\includegraphics[width=.16\linewidth,height=1.75cm]{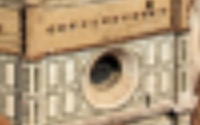}&
\includegraphics[width=.16\linewidth,height=1.75cm]{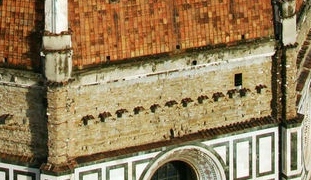}&
\includegraphics[width=.16\linewidth,height=1.75cm]{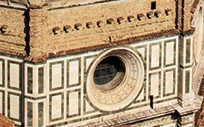}&
\includegraphics[width=.16\linewidth,height=1.75cm]{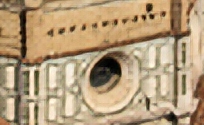} &
\includegraphics[width=.16\linewidth,height=1.75cm]{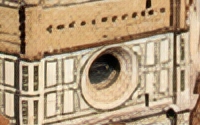} \\ 
\includegraphics[width=.16\linewidth,height=1.75cm]{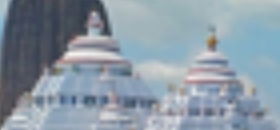}&
\includegraphics[width=.16\linewidth,height=1.75cm]{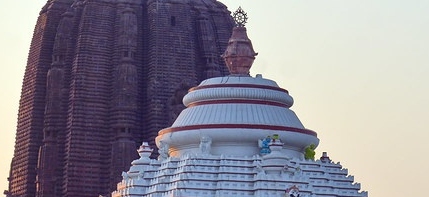}&
\includegraphics[width=.16\linewidth,height=1.75cm]{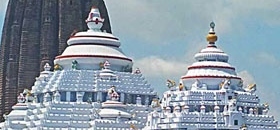}&
\includegraphics[width=.16\linewidth,height=1.75cm]{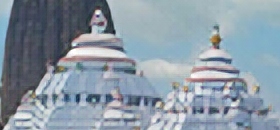} &
\includegraphics[width=.16\linewidth,height=1.75cm]{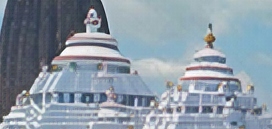} \\ 
\end{tabular}
\caption{Qualitative comparisons from ablation studies: (a) with all components intact, and (j) after excluding SE, PAR, and LSCs modules, evaluated on CUFED and Webly-Referenced SR datasets.}
\label{visual_ablation}
\end{figure}
\subsection{Further Analysis}

\textbf{Ablation studies.} We conducted ten ablative studies, as detailed in Table~\ref{ablation_study}. The proposed HiTSR incorporates several components, including the SE module, PAR module, and long skip connections (LSCs), integrated with self- and cross-attentions through a gating mechanism. These studies systematically isolate individual components within the framework to assess their impact. Beginning with the full architecture, we evaluated the effectiveness of self-attention, cross-attention modules, and gating attention. Since study (a) demonstrates superior performance compared to other ablative studies, we retain self- and cross-attention within the framework and proceed to ablate other components to evaluate the effectiveness of additional modules. Altering the architecture resulted in a performance decrease of $4.4\%$ in SSIM and $4\%$ in PSNR values, as observed in the comparison between ablation studies (a) and (j). Figure~\ref{visual_ablation} visually compares the impact of the integrated modules in our framework when functioning collectively and the effects when individual elements—SE, PAR, and LSCs—are removed.
\begin{figure}[thb!]
\centering
\begin{tabular}{cc}
\multicolumn{1}{c}{\includegraphics[scale=.45]{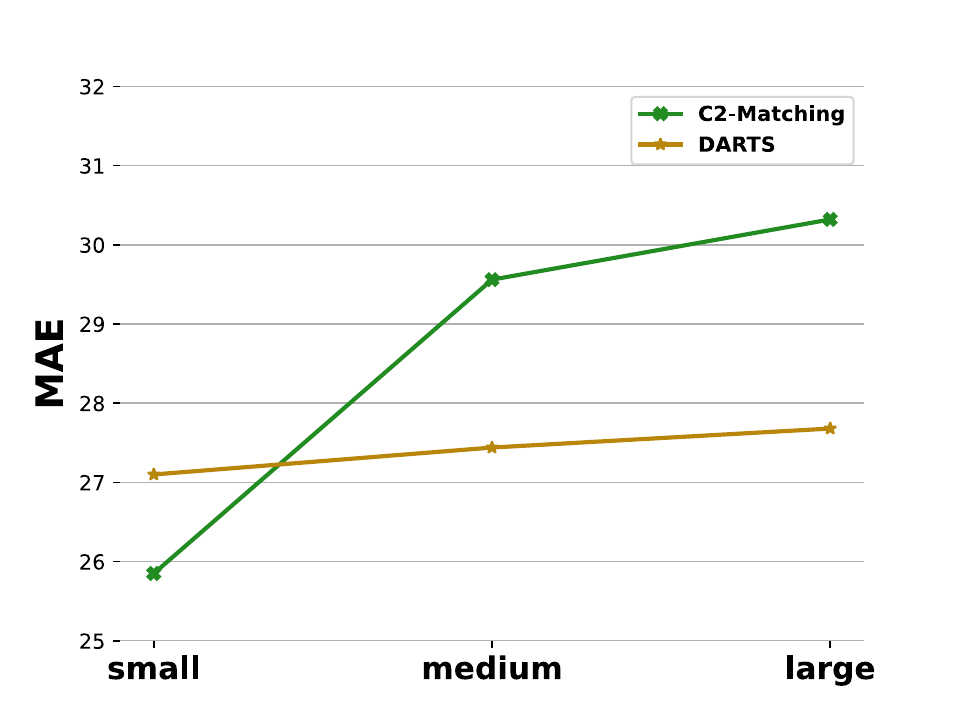}}&
\multicolumn{1}{c}{\includegraphics[scale=.45]{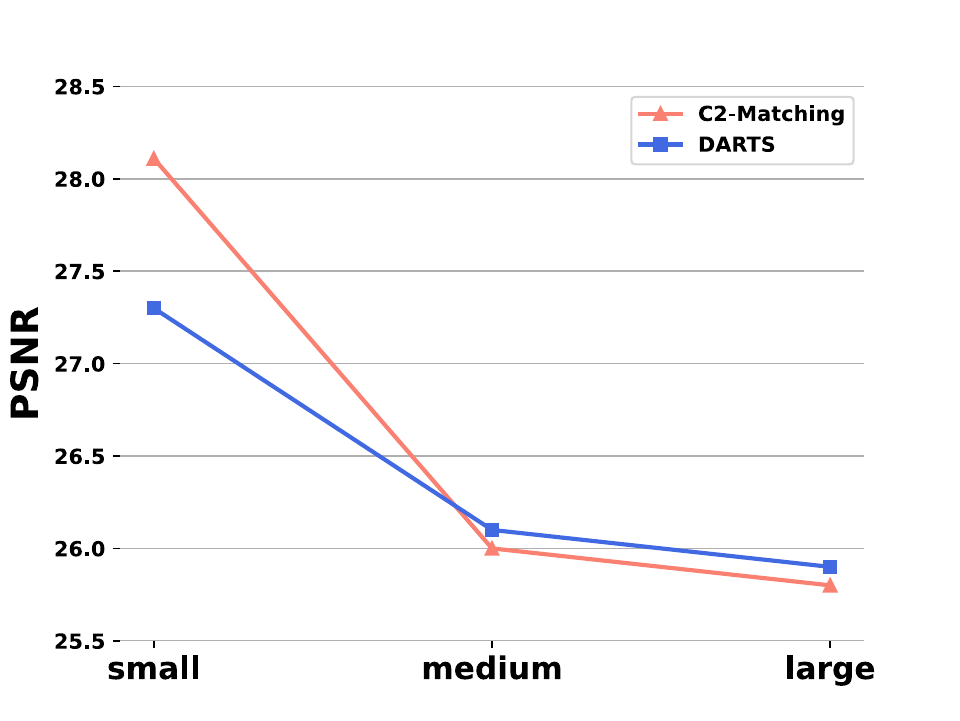}}\\
\multicolumn{2}{c}{a) Robustness to scale transformation} \\
\multicolumn{1}{c}{\includegraphics[scale=.45]{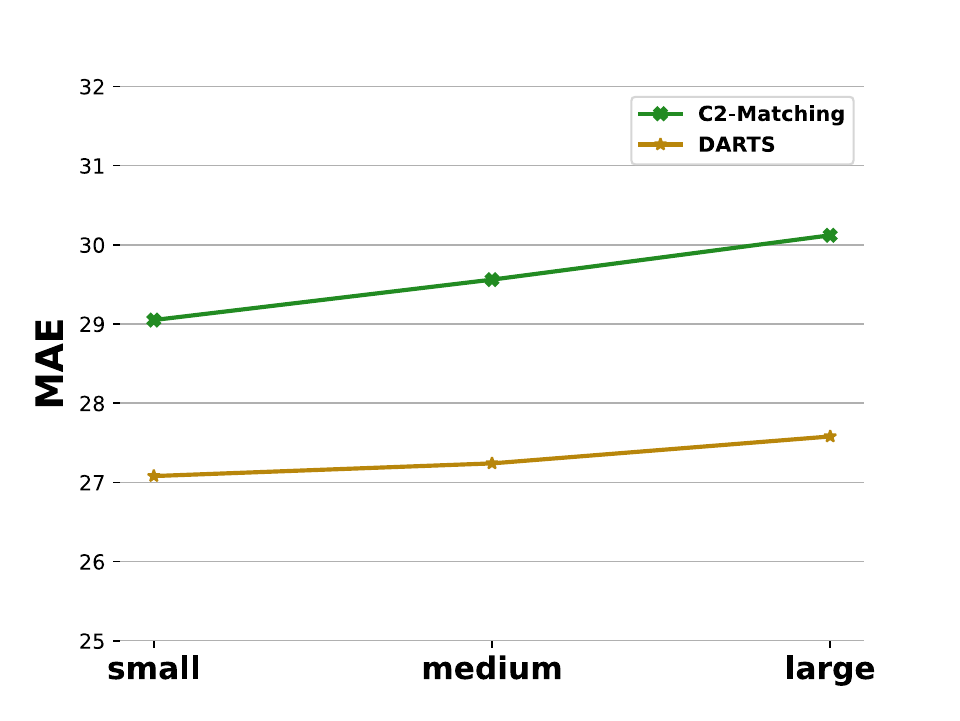}}&
\multicolumn{1}{c}{\includegraphics[scale=.45]{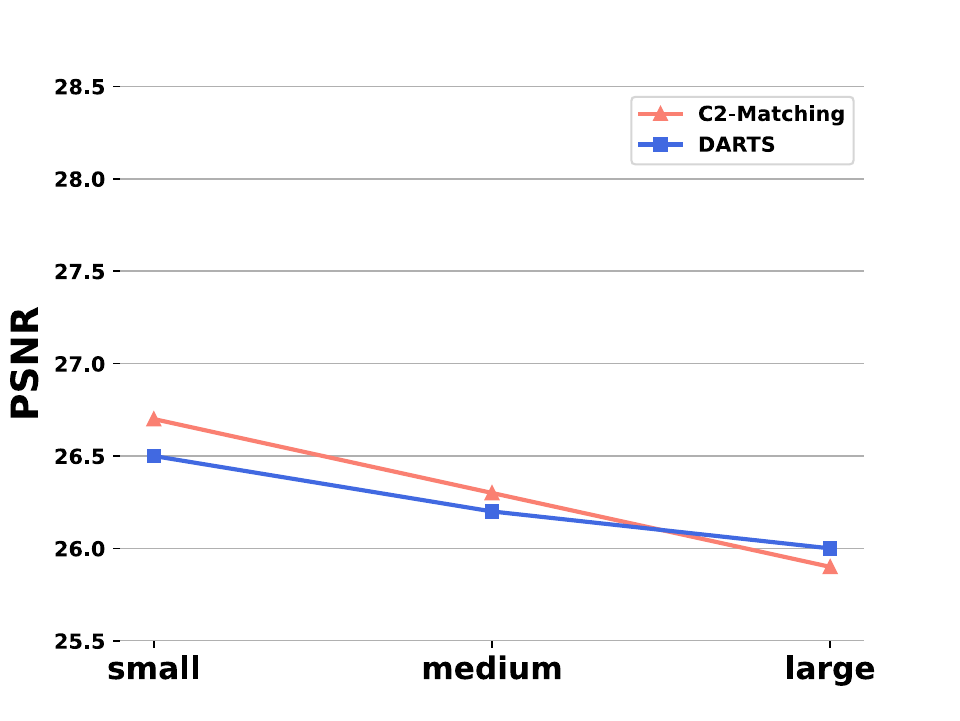}}\\
\multicolumn{2}{c}{b) Robustness to rotation transformation}
\end{tabular}
\caption{HiTSR demonstrates robustness to scale and rotation variations during inference compared to $C^2$-Matching \cite{jiang2021robust}, particularly in contour detection. The resilience to scale transformation is depicted in (a) and to rotation transformation in (b), across three augmentation levels (small, medium, large), evaluated using Mean Absolute Error (MAE, lower is better) and Peak Signal-to-Noise Ratio (PSNR, higher is better).}
\label{graphs}
\end{figure}

\textbf{Scale and Rotation Invariance.} To assess the robustness of our model to scale and rotation variations, we conduct an analysis akin to the investigations performed in \cite{jiang2021robust}. A modified CUFED5 dataset introduces diverse scale and rotation transformations—small, medium, and large alterations. Scaled and rotated input images exclusively serve as references during inference, with scaling factors from $25\%$ to $75\%$  for small to large rescaling and rotation angles uniformly distributed from 45 degrees to 135 degrees to represent small to large rotations, resulting in warped HR reference images. Our evaluation strategy employs Mean Absolute Error (MAE) across detected contours and PSNR for each image pair. This assessment enables us to quantify the accuracy of shape reconstruction and image restoration across various augmentations. Figure \ref{graphs} illustrates HiTSR' robustness to scale and rotation rotations during inference, although $C^2$-Matching outperforms HiTSR in PSNR when less variation is applied to reference data. With increasing transformation degrees, $C^2$-Matching exhibits more mismatched reconstructed features than HiTSR. We did not use large transformations during training. According to PSNR, HiTSR shows slightly more robust restoration performance than $C^2$-Matching in medium and small transformations.
\subsection{Model Size and Training Stages Comparison}
\label{sec:supp_B}
The comparison of model sizes, reflecting the number of training parameters, is detailed in Table \ref{params}. The presented HiTSR model features 13.72 million trainable parameters, incorporating double attention across two image distributions and combining 6 transformer blocks, each with a sequence of two depths and four heads. The analysis of all models is specifically conducted using only the $l_1$ loss function, and no discriminator is incorporated in the process. Additionally, Table \ref{params} compares the number of subnetworks for training each model.

$C^2$-Matching and its variants, such as DATSR \cite{cao2022reference} and RRSR \cite{zhang2022rrsr}, consist of multiple subnetworks. In Table \ref{params}, only the number of parameters of their third trained subnetwork is reported. DATSR \cite{cao2022reference} integrates transformer blocks within this subnetwork. Our model, HiTSR, with its streamlined architecture and fewer parameters, outperformed DATSR in three benchmarks — SUN80, Urban100, and Manga109 — and achieved competitive results in the CUFED5 and Webly-referenced SR datasets.

\begin{table}[ht!]
\centering
\begin{tabular}{ c|c|c} 
\toprule
Methods & \#Params & \# Training subnetworks\\
\midrule
SRNTT-\textit{rec} \cite{zhang2019image}& 5.75M & 1 \\ 
TTSR-\textit{rec} \cite{yang2020learning}& 6.2M  & 1\\ 
$C2$-Matching-\textit{rec} \cite{jiang2021robust}& 8.9M  & 3\\ 
DATSR-\textit{rec} \cite{cao2022reference} & 18.0M & 3\\ 
HiTSR-\textit{rec} & 13.72M & 1\\ 
\bottomrule
\end{tabular}
\caption{Analyzing network parameters: The '-rec' suffix indicates the inclusion of only the reconstruction ($l_1$-norm) loss. Our model size is influenced by four windows spanning two image distributions across a total of 6 transformer blocks.}
\label{params}
\end{table}
\section{Conclusion}
\label{sec:conclusion}
This study presents HiTSR, a hierarchical transformer model for reference-based image super-resolution. HiTSR features a two-stream architecture with self-attention and cross-attention, utilizing a gating-attention strategy. The model efficiently captures global context by incorporating global tokens and interacting with features in the low-resolution input image's distribution. Long skip connections facilitate efficient communication between shallow and deep layers. The architecture is conceptually straightforward, with a single end-to-end trainable module and a single training stage. Quantitative and qualitative evaluations demonstrate competitive performance compared to prior state-of-the-art approaches, often involving complex, multi-stage processes. HiTSR also exhibits reduced sensitivity to scale and rotation transformations in reference images. Introducing a novel direction for reference-based image super-resolution, HiTSR emphasizes the consolidation of the attention mechanism as an alternative to more intricate methods like $C^2$-Matching.
\section*{Conflict of Interest Statement}
The authors declare that the research was conducted without any commercial or financial relationships that could potentially create a conflict of interest.

\section*{Author Contributions}
MA implemented the software and experiments. MA, JU, and IS wrote the manuscript jointly.
\medskip
{\small
\bibliographystyle{abbrvnat}
\bibliography{neurips_2023.bib}

\begin{thebibliography}{54}
\providecommand{\natexlab}[1]{#1}
\providecommand{\url}[1]{\texttt{#1}}
\expandafter\ifx\csname urlstyle\endcsname\relax
  \providecommand{\doi}[1]{doi: #1}\else
  \providecommand{\doi}{doi: \begingroup \urlstyle{rm}\Url}\fi

\bibitem[Cao et~al.(2022)Cao, Liang, Zhang, Li, Zhang, Wang, and
  Gool]{cao2022reference}
J.~Cao, J.~Liang, K.~Zhang, Y.~Li, Y.~Zhang, W.~Wang, and L.~V. Gool.
\newblock Reference-based image super-resolution with deformable attention
  transformer.
\newblock In \emph{Computer Vision--ECCV 2022: 17th European Conference, Tel
  Aviv, Israel, October 23--27, 2022, Proceedings, Part XVIII}, pages 325--342.
  Springer, 2022.

\bibitem[Chen et~al.(2023)Chen, Wang, Zhou, Qiao, and Dong]{chen2023activating}
X.~Chen, X.~Wang, J.~Zhou, Y.~Qiao, and C.~Dong.
\newblock Activating more pixels in image super-resolution transformer.
\newblock In \emph{Proceedings of the IEEE/CVF conference on computer vision
  and pattern recognition}, pages 22367--22377, 2023.

\bibitem[Choi et~al.(2021)Choi, Lee, Jeong, and Yoon]{choi2021toward}
J.~Choi, J.~Lee, Y.~Jeong, and S.~Yoon.
\newblock Toward spatially unbiased generative models.
\newblock \emph{arXiv preprint arXiv:2108.01285}, 2021.

\bibitem[Conde et~al.(2023)Conde, Choi, Burchi, and Timofte]{conde2023swin2sr}
M.~V. Conde, U.-J. Choi, M.~Burchi, and R.~Timofte.
\newblock Swin2sr: Swinv2 transformer for compressed image super-resolution and
  restoration.
\newblock In \emph{Computer Vision--ECCV 2022 Workshops: Tel Aviv, Israel,
  October 23--27, 2022, Proceedings, Part II}, pages 669--687. Springer, 2023.

\bibitem[Dai et~al.(2019)Dai, Cai, Zhang, Xia, and Zhang]{dai2019second}
T.~Dai, J.~Cai, Y.~Zhang, S.-T. Xia, and L.~Zhang.
\newblock Second-order attention network for single image super-resolution.
\newblock In \emph{Proceedings of the IEEE/CVF conference on computer vision
  and pattern recognition}, pages 11065--11074, 2019.

\bibitem[d'Ascoli et~al.(2021)d'Ascoli, Touvron, Leavitt, Morcos, Biroli, and
  Sagun]{d2021convit}
S.~d'Ascoli, H.~Touvron, M.~Leavitt, A.~Morcos, G.~Biroli, and L.~Sagun.
\newblock Convit: Improving vision transformers with soft convolutional
  inductive biases.
\newblock \emph{arXiv preprint arXiv:2103.10697}, 2021.

\bibitem[Dong et~al.(2015)Dong, Loy, He, and Tang]{dong2015image}
C.~Dong, C.~C. Loy, K.~He, and X.~Tang.
\newblock Image super-resolution using deep convolutional networks.
\newblock \emph{IEEE transactions on pattern analysis and machine
  intelligence}, 38\penalty0 (2):\penalty0 295--307, 2015.

\bibitem[Dong et~al.(2016)Dong, Loy, and Tang]{dong2016accelerating}
C.~Dong, C.~C. Loy, and X.~Tang.
\newblock Accelerating the super-resolution convolutional neural network.
\newblock In \emph{European conference on computer vision}, pages 391--407.
  Springer, 2016.

\bibitem[Dong et~al.(2024)Dong, Yuan, Luo, Chen, Zhang, Zhang, Li, Zheng, and
  Fu]{dong2024building}
R.~Dong, S.~Yuan, B.~Luo, M.~Chen, J.~Zhang, L.~Zhang, W.~Li, J.~Zheng, and
  H.~Fu.
\newblock Building bridges across spatial and temporal resolutions:
  Reference-based super-resolution via change priors and conditional diffusion
  model.
\newblock \emph{arXiv preprint arXiv:2403.17460}, 2024.

\bibitem[Goodfellow et~al.(2020)Goodfellow, Pouget-Abadie, Mirza, Xu,
  Warde-Farley, Ozair, Courville, and Bengio]{goodfellow2020generative}
I.~Goodfellow, J.~Pouget-Abadie, M.~Mirza, B.~Xu, D.~Warde-Farley, S.~Ozair,
  A.~Courville, and Y.~Bengio.
\newblock Generative adversarial networks.
\newblock \emph{Communications of the ACM}, 63\penalty0 (11):\penalty0
  139--144, 2020.

\bibitem[Hatamizadeh et~al.(2023)Hatamizadeh, Yin, Heinrich, Kautz, and
  Molchanov]{hatamizadeh2023global}
A.~Hatamizadeh, H.~Yin, G.~Heinrich, J.~Kautz, and P.~Molchanov.
\newblock Global context vision transformers.
\newblock In \emph{International Conference on Machine Learning}, pages
  12633--12646. PMLR, 2023.

\bibitem[Hu et~al.(2018)Hu, Shen, and Sun]{hu2018squeeze}
J.~Hu, L.~Shen, and G.~Sun.
\newblock Squeeze-and-excitation networks.
\newblock In \emph{Proceedings of the IEEE conference on computer vision and
  pattern recognition}, pages 7132--7141, 2018.

\bibitem[Huang et~al.(2015)Huang, Singh, and Ahuja]{huang2015single}
J.-B. Huang, A.~Singh, and N.~Ahuja.
\newblock Single image super-resolution from transformed self-exemplars.
\newblock In \emph{Proceedings of the IEEE conference on computer vision and
  pattern recognition}, pages 5197--5206, 2015.

\bibitem[Jiang et~al.(2021)Jiang, Chan, Wang, Loy, and Liu]{jiang2021robust}
Y.~Jiang, K.~C. Chan, X.~Wang, C.~C. Loy, and Z.~Liu.
\newblock Robust reference-based super-resolution via c2-matching.
\newblock In \emph{Proceedings of the IEEE/CVF Conference on Computer Vision
  and Pattern Recognition}, pages 2103--2112, 2021.

\bibitem[Johnson et~al.(2016)Johnson, Alahi, and
  Fei-Fei]{johnson2016perceptual}
J.~Johnson, A.~Alahi, and L.~Fei-Fei.
\newblock Perceptual losses for real-time style transfer and super-resolution.
\newblock In \emph{Computer Vision--ECCV 2016: 14th European Conference,
  Amsterdam, The Netherlands, October 11-14, 2016, Proceedings, Part II 14},
  pages 694--711. Springer, 2016.

\bibitem[Karras et~al.(2020)Karras, Laine, Aittala, Hellsten, Lehtinen, and
  Aila]{karras2020analyzing}
T.~Karras, S.~Laine, M.~Aittala, J.~Hellsten, J.~Lehtinen, and T.~Aila.
\newblock Analyzing and improving the image quality of stylegan.
\newblock In \emph{Proceedings of the IEEE/CVF conference on computer vision
  and pattern recognition}, pages 8110--8119, 2020.

\bibitem[Kim et~al.(2016{\natexlab{a}})Kim, Lee, and Lee]{kim2016accurate}
J.~Kim, J.~K. Lee, and K.~M. Lee.
\newblock Accurate image super-resolution using very deep convolutional
  networks.
\newblock In \emph{Proceedings of the IEEE conference on computer vision and
  pattern recognition}, pages 1646--1654, 2016{\natexlab{a}}.

\bibitem[Kim et~al.(2016{\natexlab{b}})Kim, Lee, and Lee]{kim2016deeply}
J.~Kim, J.~K. Lee, and K.~M. Lee.
\newblock Deeply-recursive convolutional network for image super-resolution.
\newblock In \emph{Proceedings of the IEEE conference on computer vision and
  pattern recognition}, pages 1637--1645, 2016{\natexlab{b}}.

\bibitem[Kingma and Ba(2014)]{kingma2014adam}
D.~P. Kingma and J.~Ba.
\newblock Adam: A method for stochastic optimization.
\newblock \emph{arXiv preprint arXiv:1412.6980}, 2014.

\bibitem[Ledig et~al.(2017)Ledig, Theis, Husz{\'a}r, Caballero, Cunningham,
  Acosta, Aitken, Tejani, Totz, Wang, et~al.]{ledig2017photo}
C.~Ledig, L.~Theis, F.~Husz{\'a}r, J.~Caballero, A.~Cunningham, A.~Acosta,
  A.~Aitken, A.~Tejani, J.~Totz, Z.~Wang, et~al.
\newblock Photo-realistic single image super-resolution using a generative
  adversarial network.
\newblock In \emph{Proceedings of the IEEE conference on computer vision and
  pattern recognition}, pages 4681--4690, 2017.

\bibitem[Li et~al.(2024)Li, Ying, Pan, Fan, and Shi]{li2024estgn}
Q.~Li, Z.~Ying, D.~Pan, Z.~Fan, and P.~Shi.
\newblock Estgn: Enhanced self-mined text guided super-resolution network for
  superior image super resolution.
\newblock In \emph{ICASSP 2024-2024 IEEE International Conference on Acoustics,
  Speech and Signal Processing (ICASSP)}, pages 3655--3659. IEEE, 2024.

\bibitem[Li et~al.(2021)Li, Sixou, and Peyrin]{li2021review}
Y.~Li, B.~Sixou, and F.~Peyrin.
\newblock A review of the deep learning methods for medical images super
  resolution problems.
\newblock \emph{Irbm}, 42\penalty0 (2):\penalty0 120--133, 2021.

\bibitem[Liang et~al.(2021)Liang, Cao, Sun, Zhang, Van~Gool, and
  Timofte]{liang2021swinir}
J.~Liang, J.~Cao, G.~Sun, K.~Zhang, L.~Van~Gool, and R.~Timofte.
\newblock Swinir: Image restoration using swin transformer.
\newblock In \emph{Proceedings of the IEEE/CVF International Conference on
  Computer Vision}, pages 1833--1844, 2021.

\bibitem[Lim et~al.(2017)Lim, Son, Kim, Nah, and Mu~Lee]{lim2017enhanced}
B.~Lim, S.~Son, H.~Kim, S.~Nah, and K.~Mu~Lee.
\newblock Enhanced deep residual networks for single image super-resolution.
\newblock In \emph{Proceedings of the IEEE conference on computer vision and
  pattern recognition workshops}, pages 136--144, 2017.

\bibitem[Lim and Ye(2017)]{lim2017geometric}
J.~H. Lim and J.~C. Ye.
\newblock Geometric gan.
\newblock \emph{arXiv preprint arXiv:1705.02894}, 2017.

\bibitem[Liu et~al.(2018)Liu, Wen, Fan, Loy, and Huang]{liu2018non}
D.~Liu, B.~Wen, Y.~Fan, C.~C. Loy, and T.~S. Huang.
\newblock Non-local recurrent network for image restoration.
\newblock \emph{Advances in neural information processing systems}, 31, 2018.

\bibitem[Liu et~al.(2020)Liu, Zhang, Tang, Tang, and Wu]{liu2020residual}
J.~Liu, W.~Zhang, Y.~Tang, J.~Tang, and G.~Wu.
\newblock Residual feature aggregation network for image super-resolution.
\newblock In \emph{Proceedings of the IEEE/CVF conference on computer vision
  and pattern recognition}, pages 2359--2368, 2020.

\bibitem[Liu et~al.(2021)Liu, Lin, Cao, Hu, Wei, Zhang, Lin, and
  Guo]{liu2021swin}
Z.~Liu, Y.~Lin, Y.~Cao, H.~Hu, Y.~Wei, Z.~Zhang, S.~Lin, and B.~Guo.
\newblock Swin transformer: Hierarchical vision transformer using shifted
  windows.
\newblock In \emph{Proceedings of the IEEE/CVF international conference on
  computer vision}, pages 10012--10022, 2021.

\bibitem[Lu et~al.(2019)Lu, Batra, Parikh, and Lee]{lu2019vilbert}
J.~Lu, D.~Batra, D.~Parikh, and S.~Lee.
\newblock Vilbert: Pretraining task-agnostic visiolinguistic representations
  for vision-and-language tasks.
\newblock In \emph{Advances in Neural Information Processing Systems}, pages
  13--23, 2019.

\bibitem[Lu et~al.(2021)Lu, Li, Tao, Lu, and Jia]{lu2021masa}
L.~Lu, W.~Li, X.~Tao, J.~Lu, and J.~Jia.
\newblock Masa-sr: Matching acceleration and spatial adaptation for
  reference-based image super-resolution.
\newblock In \emph{Proceedings of the IEEE/CVF Conference on Computer Vision
  and Pattern Recognition}, pages 6368--6377, 2021.

\bibitem[Matsui et~al.(2017)Matsui, Ito, Aramaki, Fujimoto, Ogawa, Yamasaki,
  and Aizawa]{matsui2017sketch}
Y.~Matsui, K.~Ito, Y.~Aramaki, A.~Fujimoto, T.~Ogawa, T.~Yamasaki, and
  K.~Aizawa.
\newblock Sketch-based manga retrieval using manga109 dataset.
\newblock \emph{Multimedia Tools and Applications}, 76:\penalty0 21811--21838,
  2017.

\bibitem[Mei et~al.(2020)Mei, Fan, Zhou, Huang, Huang, and Shi]{mei2020image}
Y.~Mei, Y.~Fan, Y.~Zhou, L.~Huang, T.~S. Huang, and H.~Shi.
\newblock Image super-resolution with cross-scale non-local attention and
  exhaustive self-exemplars mining.
\newblock In \emph{Proceedings of the IEEE/CVF conference on computer vision
  and pattern recognition}, pages 5690--5699, 2020.

\bibitem[Mei et~al.(2021)Mei, Fan, and Zhou]{mei2021image}
Y.~Mei, Y.~Fan, and Y.~Zhou.
\newblock Image super-resolution with non-local sparse attention.
\newblock In \emph{Proceedings of the IEEE/CVF Conference on Computer Vision
  and Pattern Recognition}, pages 3517--3526, 2021.

\bibitem[Ronneberger et~al.(2015)Ronneberger, Fischer, and
  Brox]{ronneberger2015u}
O.~Ronneberger, P.~Fischer, and T.~Brox.
\newblock U-net: Convolutional networks for biomedical image segmentation.
\newblock In \emph{Medical Image Computing and Computer-Assisted
  Intervention--MICCAI 2015: 18th International Conference, Munich, Germany,
  October 5-9, 2015, Proceedings, Part III 18}, pages 234--241. Springer, 2015.

\bibitem[Sajjadi et~al.(2017)Sajjadi, Scholkopf, and
  Hirsch]{sajjadi2017enhancenet}
M.~S. Sajjadi, B.~Scholkopf, and M.~Hirsch.
\newblock Enhancenet: Single image super-resolution through automated texture
  synthesis.
\newblock In \emph{Proceedings of the IEEE international conference on computer
  vision}, pages 4491--4500, 2017.

\bibitem[Shi et~al.(2016)Shi, Caballero, Husz{\'a}r, Totz, Aitken, Bishop,
  Rueckert, and Wang]{shi2016real}
W.~Shi, J.~Caballero, F.~Husz{\'a}r, J.~Totz, A.~P. Aitken, R.~Bishop,
  D.~Rueckert, and Z.~Wang.
\newblock Real-time single image and video super-resolution using an efficient
  sub-pixel convolutional neural network.
\newblock In \emph{Proceedings of the IEEE conference on computer vision and
  pattern recognition}, pages 1874--1883, 2016.

\bibitem[Simonyan and Zisserman(2014)]{simonyan2014very}
K.~Simonyan and A.~Zisserman.
\newblock Very deep convolutional networks for large-scale image recognition.
\newblock \emph{arXiv preprint arXiv:1409.1556}, 2014.

\bibitem[Sun and Hays(2012)]{sun2012super}
L.~Sun and J.~Hays.
\newblock Super-resolution from internet-scale scene matching.
\newblock In \emph{2012 IEEE International conference on computational
  photography (ICCP)}, pages 1--12. IEEE, 2012.

\bibitem[Tong et~al.(2017)Tong, Li, Liu, and Gao]{tong2017image}
T.~Tong, G.~Li, X.~Liu, and Q.~Gao.
\newblock Image super-resolution using dense skip connections.
\newblock In \emph{Proceedings of the IEEE international conference on computer
  vision}, pages 4799--4807, 2017.

\bibitem[Vaswani et~al.(2017)Vaswani, Shazeer, Parmar, Uszkoreit, Jones, Gomez,
  Kaiser, and Polosukhin]{vaswani2017attention}
A.~Vaswani, N.~Shazeer, N.~Parmar, J.~Uszkoreit, L.~Jones, A.~N. Gomez,
  {\L}.~Kaiser, and I.~Polosukhin.
\newblock Attention is all you need.
\newblock \emph{Advances in neural information processing systems}, 30, 2017.

\bibitem[Wang et~al.(2018{\natexlab{a}})Wang, Liu, Zhu, Tao, Kautz, and
  Catanzaro]{wang2018high}
T.-C. Wang, M.-Y. Liu, J.-Y. Zhu, A.~Tao, J.~Kautz, and B.~Catanzaro.
\newblock High-resolution image synthesis and semantic manipulation with
  conditional gans.
\newblock In \emph{Proceedings of the IEEE conference on computer vision and
  pattern recognition}, pages 8798--8807, 2018{\natexlab{a}}.

\bibitem[Wang et~al.(2018{\natexlab{b}})Wang, Yu, Wu, Gu, Liu, Dong, Qiao, and
  Change~Loy]{wang2018esrgan}
X.~Wang, K.~Yu, S.~Wu, J.~Gu, Y.~Liu, C.~Dong, Y.~Qiao, and C.~Change~Loy.
\newblock Esrgan: Enhanced super-resolution generative adversarial networks.
\newblock In \emph{Proceedings of the European conference on computer vision
  (ECCV) workshops}, pages 0--0, 2018{\natexlab{b}}.

\bibitem[Wang et~al.(2004)Wang, Bovik, Sheikh, and Simoncelli]{wang2004image}
Z.~Wang, A.~C. Bovik, H.~R. Sheikh, and E.~P. Simoncelli.
\newblock Image quality assessment: from error visibility to structural
  similarity.
\newblock \emph{IEEE transactions on image processing}, 13\penalty0
  (4):\penalty0 600--612, 2004.

\bibitem[Xu et~al.(2021)Xu, Wang, Chen, Zhou, and Loy]{xu2021positional}
R.~Xu, X.~Wang, K.~Chen, B.~Zhou, and C.~C. Loy.
\newblock Positional encoding as spatial inductive bias in gans.
\newblock In \emph{Proceedings of the IEEE/CVF Conference on Computer Vision
  and Pattern Recognition}, pages 13569--13578, 2021.

\bibitem[Yang et~al.(2020)Yang, Yang, Fu, Lu, and Guo]{yang2020learning}
F.~Yang, H.~Yang, J.~Fu, H.~Lu, and B.~Guo.
\newblock Learning texture transformer network for image super-resolution.
\newblock In \emph{Proceedings of the IEEE/CVF conference on computer vision
  and pattern recognition}, pages 5791--5800, 2020.

\bibitem[Zhang et~al.(2022{\natexlab{a}})Zhang, Gu, Zhang, Bao, Chen, Wen,
  Wang, and Guo]{zhang2022styleswin}
B.~Zhang, S.~Gu, B.~Zhang, J.~Bao, D.~Chen, F.~Wen, Y.~Wang, and B.~Guo.
\newblock Styleswin: Transformer-based gan for high-resolution image
  generation.
\newblock In \emph{Proceedings of the IEEE/CVF conference on computer vision
  and pattern recognition}, pages 11304--11314, 2022{\natexlab{a}}.

\bibitem[Zhang et~al.(2022{\natexlab{b}})Zhang, Huang, Liu, Wang, and
  Jin]{zhang2022swinfir}
D.~Zhang, F.~Huang, S.~Liu, X.~Wang, and Z.~Jin.
\newblock Swinfir: Revisiting the swinir with fast fourier convolution and
  improved training for image super-resolution.
\newblock \emph{arXiv preprint arXiv:2208.11247}, 2022{\natexlab{b}}.

\bibitem[Zhang et~al.(2022{\natexlab{c}})Zhang, Zhang, Gu, Zhang, Kong, and
  Yuan]{zhang2022accurate}
J.~Zhang, Y.~Zhang, J.~Gu, Y.~Zhang, L.~Kong, and X.~Yuan.
\newblock Accurate image restoration with attention retractable transformer.
\newblock \emph{arXiv preprint arXiv:2210.01427}, 2022{\natexlab{c}}.

\bibitem[Zhang et~al.(2022{\natexlab{d}})Zhang, Li, He, Li, Wang, and
  Zhang]{zhang2022rrsr}
L.~Zhang, X.~Li, D.~He, F.~Li, Y.~Wang, and Z.~Zhang.
\newblock Rrsr: Reciprocal reference-based image super-resolution with
  progressive feature alignment and selection.
\newblock In \emph{Computer Vision--ECCV 2022: 17th European Conference, Tel
  Aviv, Israel, October 23--27, 2022, Proceedings, Part XIX}, pages 648--664.
  Springer, 2022{\natexlab{d}}.

\bibitem[Zhang et~al.(2019{\natexlab{a}})Zhang, Liu, Dong, and
  Qiao]{zhang2019ranksrgan}
W.~Zhang, Y.~Liu, C.~Dong, and Y.~Qiao.
\newblock Ranksrgan: Generative adversarial networks with ranker for image
  super-resolution.
\newblock In \emph{Proceedings of the IEEE/CVF International Conference on
  Computer Vision}, pages 3096--3105, 2019{\natexlab{a}}.

\bibitem[Zhang et~al.(2018)Zhang, Li, Li, Wang, Zhong, and Fu]{zhang2018image}
Y.~Zhang, K.~Li, K.~Li, L.~Wang, B.~Zhong, and Y.~Fu.
\newblock Image super-resolution using very deep residual channel attention
  networks.
\newblock In \emph{Proceedings of the European conference on computer vision
  (ECCV)}, pages 286--301, 2018.

\bibitem[Zhang et~al.(2019{\natexlab{b}})Zhang, Li, Li, Zhong, and
  Fu]{zhang2019residual}
Y.~Zhang, K.~Li, K.~Li, B.~Zhong, and Y.~Fu.
\newblock Residual non-local attention networks for image restoration.
\newblock \emph{arXiv preprint arXiv:1903.10082}, 2019{\natexlab{b}}.

\bibitem[Zhang et~al.(2019{\natexlab{c}})Zhang, Wang, Lin, and
  Qi]{zhang2019image}
Z.~Zhang, Z.~Wang, Z.~Lin, and H.~Qi.
\newblock Image super-resolution by neural texture transfer.
\newblock In \emph{Proceedings of the IEEE/CVF conference on computer vision
  and pattern recognition}, pages 7982--7991, 2019{\natexlab{c}}.

\bibitem[Zheng et~al.(2018)Zheng, Ji, Wang, Liu, and Fang]{zheng2018crossnet}
H.~Zheng, M.~Ji, H.~Wang, Y.~Liu, and L.~Fang.
\newblock Crossnet: An end-to-end reference-based super resolution network
  using cross-scale warping.
\newblock In \emph{Proceedings of the European conference on computer vision
  (ECCV)}, pages 88--104, 2018.

\end{thebibliography}
}




\end{document}